\pdfoutput=1

\documentclass[11pt]{article}

\usepackage[final]{acl}

\usepackage{times}
\usepackage{latexsym}

\usepackage[T1]{fontenc}

\usepackage[utf8]{inputenc}

\usepackage{microtype}

\usepackage{inconsolata}

\usepackage{graphicx}
\usepackage{array}
\usepackage{makecell}
\usepackage{bm}
\usepackage{multirow}
\usepackage{amsfonts}
\usepackage{xcolor} 
\usepackage{colortbl} 
\usepackage{booktabs}
\usepackage{url}
\usepackage[normalem]{ulem}

\newcommand\blfootnote[1]{%
  \begingroup
  \renewcommand\thefootnote{}\footnote{#1}%
  \addtocounter{footnote}{-1}%
  \endgroup
}

\title{TableDreamer: Progressive and Weakness-guided Data Synthesis from Scratch for Table Instruction Tuning}

\author{Mingyu Zheng$^{1,2\dagger}$,\ Zhifan Feng$^{3}$,\ Jia Wang$^{1,2}$,\ Lanrui Wang$^{1,2}$, \\ {\bf Zheng Lin$^{1,2\ddagger}$, Yang Hao$^3$, Weiping Wang$^{1}$}  \\
$^1$Institute of Information Engineering, Chinese Academy of Sciences, Beijing, China \\
$^2$School of Cyber Security, University of Chinese Academy of Sciences, Beijing, China \\
$^3$Baidu Inc, Beijing, China \\
\texttt{\{zhengmingyu,wangjia,wanglanrui,linzheng,wangweiping\}@iie.ac.cn}\\
\texttt{\{fengzhifan,haoyang03\}@baidu.com}\\
}

\begin{document}
\maketitle
\begin{abstract}
Despite the commendable progress of recent LLM-based data synthesis methods, they face two limitations in generating table instruction tuning data. First, they can not thoroughly explore the vast input space of table understanding tasks, leading to limited data diversity. Second, they ignore the weaknesses in table understanding ability of the target LLM and blindly pursue the increase of data quantity, resulting in suboptimal data efficiency. In this paper, we introduce a progressive and weakness-guided data synthesis framework tailored for table instruction tuning, named TableDreamer, to mitigate the above issues. Specifically, we first synthesize diverse tables and related instructions as seed data, and then perform an iterative exploration of the input space under the guidance of the newly identified weakness data, which eventually serve as the final training data for fine-tuning the target LLM. Extensive experiments on 10 tabular benchmarks demonstrate the effectiveness of the proposed framework, which boosts the average accuracy of Llama3.1-8B-instruct by 11.62\% ($49.07\%\rightarrow60.69\%$) with 27K GPT-4o synthetic data and outperforms state-of-the-art data synthesis baselines which use more training data. The code and data is available at \url{https://github.com/SpursGoZmy/TableDreamer}.
\blfootnote{$^{\dagger}$This work was done during an internship at Baidu Inc.}
\blfootnote{$^{\ddagger}$ Corresponding author: Zheng Lin.}

\end{abstract}

\section{Introduction}

Table understanding technique aims to enable models to automatically comprehend tables and complete various table-related tasks~\cite{llm_for_table_processing_survey,tu_survey_2023}. With the recent advancement of large language models (LLMs), the dominant paradigm for table understanding has shifted to instruction tuning general LLMs with tabular task data, leading to the rise of powerful Tabular LLMs~\cite{tablellama,tablegpt_microsoft}. 

\begin{figure}[t]
  \centering
  \includegraphics[width=\linewidth]{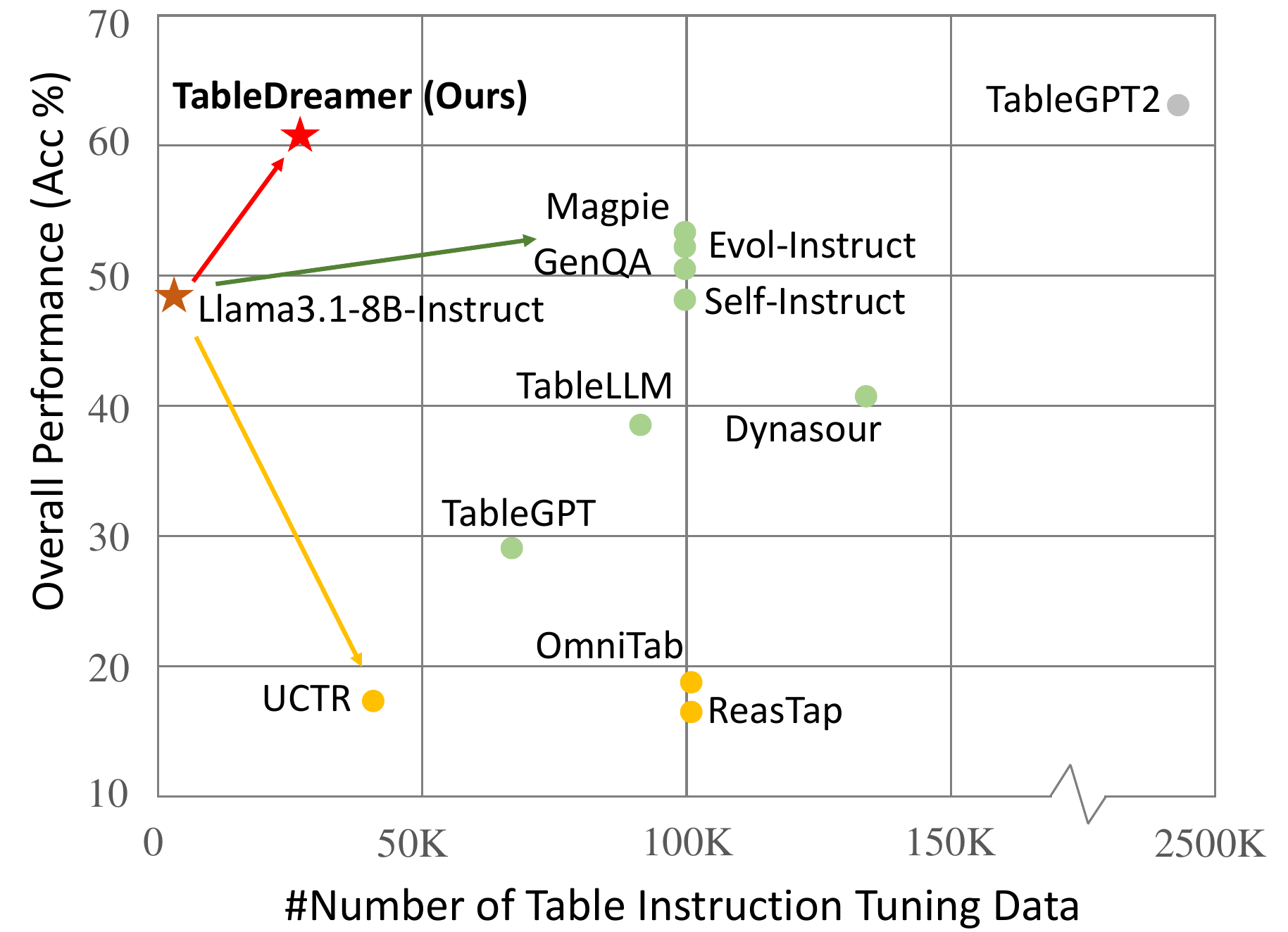}
  \caption{ The comparison of performance and training data volume between TableDreamer and previous table instruction tuning data synthesis methods over 10 tabular benchmarks.}
  \label{performance_scatter}
\end{figure}

In early work on tabular LLMs, instruction-tuning samples were manually collected by human annotators or converted from public datasets using fixed instruction templates. However, the reusing of existing datasets often leads to poor task and data diversity, while human annotation also faces the challenge of prohibitively expensive cost. Therefore, researchers turned to employ LLMs to generate table instruction tuning data. For instance, \citet{tablellm} uses GPT-3.5 to generate questions based on benchmark tables, which serve as the training data for fine-tuning CodeLlama~\cite{codellama}. The resulting TableLLM model outperforms general LLMs on several tabular benchmarks, demonstrating the potential of synthetic data in table instruction tuning.

Although existing data synthesis approaches have achieved commendable performance, they still face two limitations in generating table instruction tuning data. First, \textbf{existing data synthesis methods are unable to fully explore the vast input space composed of input tables and instructions, leading to limited data diversity.} On the one hand, general data generation methods like Self-Instruct~\cite{wang-etal-2023-self-instruct} primarily focus on generating unstructured text data, and they did not adequately consider the unique characteristics of structured tables (e.g., diverse table structures, different table formats). As a result, they tend to produce simple tables and instructions of limited tabular tasks. On the other hand, existing studies on tabular LLMs only explore how to synthesize more instructions based on directly available tables from public datasets to improve instruction diversity, but they lack the ability to synthesize more diversified tabular data, which also limits the diversity of the final table instruction tuning data.

Second, \textbf{existing data synthesis methods ignore the LLM's weaknesses in table understanding ability, resulting in suboptimal efficiency of synthetic data.} The combination of the input table and the instruction allows us to easily create a large amount of table instruction tuning data, e.g., we can utilize an LLM to generate dozens of questions based on a single table. However, published studies have indicated that merely pursuing an increase in the quantity of instruction tuning data does not necessarily yield performance improvement~\cite{zhou2023lima,alpaca_cot}. Given the vast input space for table understanding tasks, it is more efficient to synthesize valuable data points that expose the deficiencies of the target LLM, rather than blindly increase the amount of synthetic data, which may result in both a waste of training resources and a decline in model performance.

\begin{figure*}[t]
  \centering
  \includegraphics[width=\linewidth]{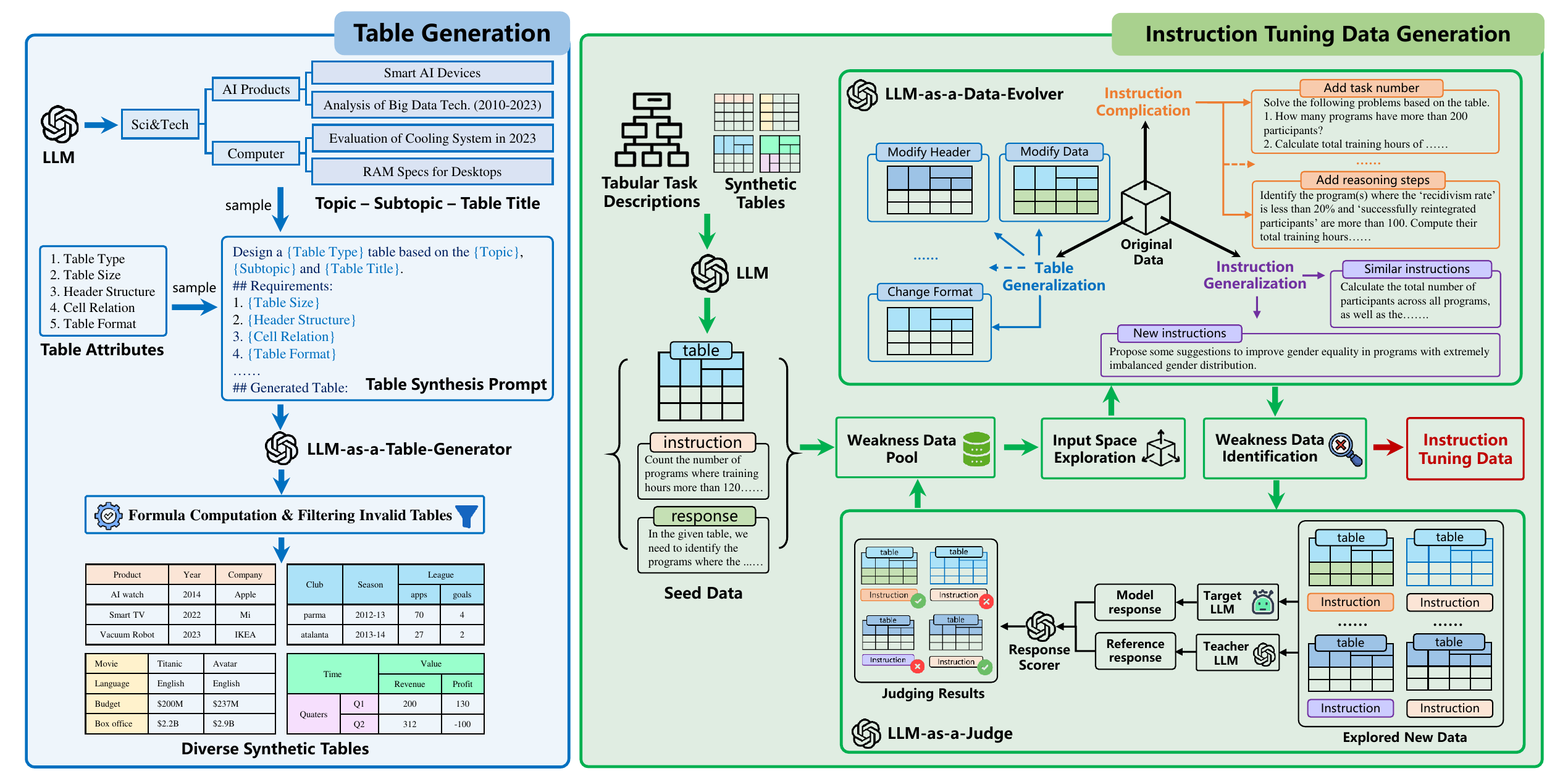}
  \caption{The overview of the proposed TableDreamer framework, which includes two stages. In stage 1, we first synthesize table titles based on different topics and subtopic, and then employ the LLM to generate diverse tables covering a wide range of key table attributes such as table structures and sizes. In stage 2, starting from a group of seed data, we perform an iterative exploration of the input space under the guidance of the newly discovered weakness data, which eventually serve as the table instruction tuning data.}
  \label{tabledreamer_framework}
\end{figure*}

To address these issues, we introduce a progressive and weakness-guided data synthesis framework for table instruction tuning, named \textbf{TableDreamer}, which can not only generate diverse tables and instructions from scratch, but can also continuously explore the input space under the guidance of newly identified weakness data to more effectively enhance the model performance. As illustrated in Figure \ref{tabledreamer_framework}, our framework consists of two stages. In stage 1, we first synthesize various table titles of different topics and subtopics, and then employ the LLM to create diverse tables. In stage 2, based on synthetic tables and tabular task descriptions, a group of seed data is generated and will undergo data evolution in three directions. The synthesized new samples are evaluated by LLM-as-a-judge to identify weakness-exposing data, which is used as the seed data for the next round of data evolution. This process can be iterated multiple times, with the accumulated weakness data serving as the final table instruction tuning data.

We compare TableDreamer with a series of data synthesis methods, general LLMs and tabular LLMs on 10 tabular benchmarks. As shown in Figure \ref{performance_scatter}, experimental results demonstrate the effectiveness of the proposed framework, which boosts the average accuracy of Llama3.1-8B-instruct by 11.62\% 
($49.07\%\rightarrow60.69\%$) with 27K GPT-4o synthetic data and outperforms the state-of-the-art data synthesis baselines that use more training data (100K+). We also demonstrate the effectiveness of TableDreamer as data augmentation in the few-shot learning scenario, where only a small number of original training samples are available (e.g., 20 samples for each benchmark). Extensive ablation experiments are conducted to reveal the contributions of different components in the framework (e.g., the influence of weakness data selection and data evolution). We hope this work could establish a strong base for future research on the table instruction tuning data synthesis and help researchers improve models' table understanding ability especially with limited annotation budget.

We conclude our contributions as follows:

1) We introduce a data synthesis framework TableDreamer tailored for table instruction tuning with better data diversity and efficiency, mitigating the limitations of current approaches.

2) We construct and release 27K table instruction tuning data, which include diverse tables and instructions of a wide range of tabular tasks that the current open-source community lacks.

3) We make a systematic investigation of existing methods to show the effectiveness of TableDreamer, which outperforms strong baselines on 10 tabular benchmarks including recent tabular LLMs.

\begin{figure}[t]
  \centering
  \includegraphics[width=0.92\linewidth]{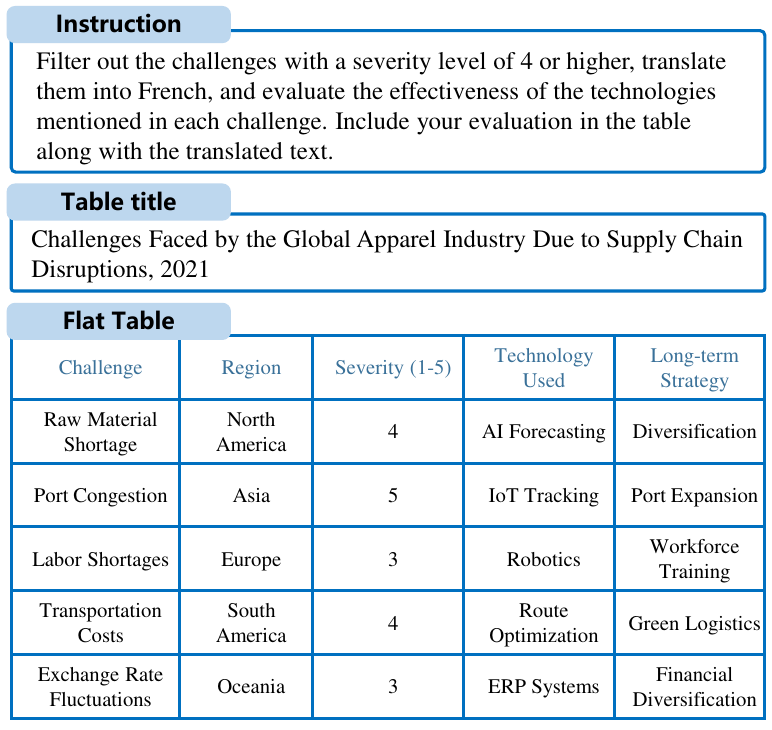}
  \caption{ Example of TableDreamer synthetic data. The synthetic table are clipped due to space limitation.}
  \label{dataset_examples_1}
\end{figure}

\section{Related Work}
\subsection{Table Instruction Tuning}

In addition to directly prompting LLMs to fulfill tabular tasks~\cite{table_cot,chain-of-table}, researchers are increasingly dedicated to developing tabular LLMs with carefully constructed table instruction tuning data. TableLlama~\cite{tablellama} collected 2.6M instruction-tuning pairs from 14 academic tabular datasets, and TableBenchLLM~\cite{tablebenchllm} even spent \$12,000 US dollars on hiring annotators for answering labeling and quality checking. Besides, LLM-based data synthesis methods were also adopted to generate table instruction tuning data. TableGPT~\cite{tablegpt_microsoft} proposed a Synthesis-then-Augment framework which uses GPT-3.5 to generate instructions based on public tables and then performs data augmentations such as instruction paraphrasing for better data diversity. TableLLM~\cite{tablellm} introduced a similar distant supervision approach which first synthesizes instructions and selects high-quality responses with the cross-way validation of different reasoning methods. However, compared with other areas like code and math, data synthesis for table instruction tuning is still in its infancy, with important issues deserving further exploration. In this paper, we introduce a novel data synthesis method, and also conduct a comprehensive investigation of relevant baselines, providing valuable insights about this emergent yet promising direction.

\subsection{LLM-based Data Synthesis}
The large amount of high-quality human-collected data has facilitated the development of deep learning in recent years. Nevertheless, purely depending on human data always involves a trade-off between data quality and quantity due to factors such as costs or privacy issues~\cite{llm_based_data_generation_survey}. Given the excellent ability to output human-like text, the advanced LLMs offer an alternative data source with synthetic data generation to mitigate drawbacks of human data. One of most prominent application of LLM-based data generation is to synthesize large-scale and diverse instruction tuning data in a cost effective way~\cite{wang-etal-2023-self-instruct,alpaca,xu2023wizardlm,glan}. Based on a handful human-created instructions as the initial seed data, Self-instruct~\cite{wang-etal-2023-self-instruct} synthesizes new instructions by prompting an LLM with randomly selected instructions from the candidate pool as few-shot demonstrations. Magpie~\cite{magpie} leverages the autoregressive nature of LLMs and elicits instructions from fine-tuned LLMs by feeding them a pre-query chat template. Unlike textual tasks, table understanding tasks poses new challenges for LLM-based data synthesis due to the hybrid input of unstructured text and structured table. Unfortunately, existing approaches usually simplify the problem setting by ignoring the demand for synthesizing diverse tables and can only generate questions using public benchmark tables. By contrast, we take a step further and explore how to synthesize both tables and relevant instructions from scratch.

\section{TableDreamer Framework}

\subsection{Problem Definition}
Given a table $T$ including its metadata like the table title and a user instruction $Inst$ about the table, the table understanding problem requires the model $f(\cdot)$ to output a response $R$ that correctly complete the specified table-related tasks in the instruction, i.e., $R=f(T,Inst)$. The goal of the table instruction tuning data synthesis is to obtain a synthetic training dataset $D_{syn}$ of $N$ triples for fine-tuning LLMs, i.e., $D_{syn} = \{ (Inst_i, T_i,R_i) \mid i = 1, 2, \ldots, N \}$. Existing data synthesis methods often simplify the problem setting by assuming that tables are always directly available, and thus only focus on generating table-related instructions.  By contrast, we retain the original setting and endeavor to synthesize diverse tables and instructions from scratch without relying on any public datasets.

\begin{figure}[t]
  \centering
  \includegraphics[width=0.95\linewidth]{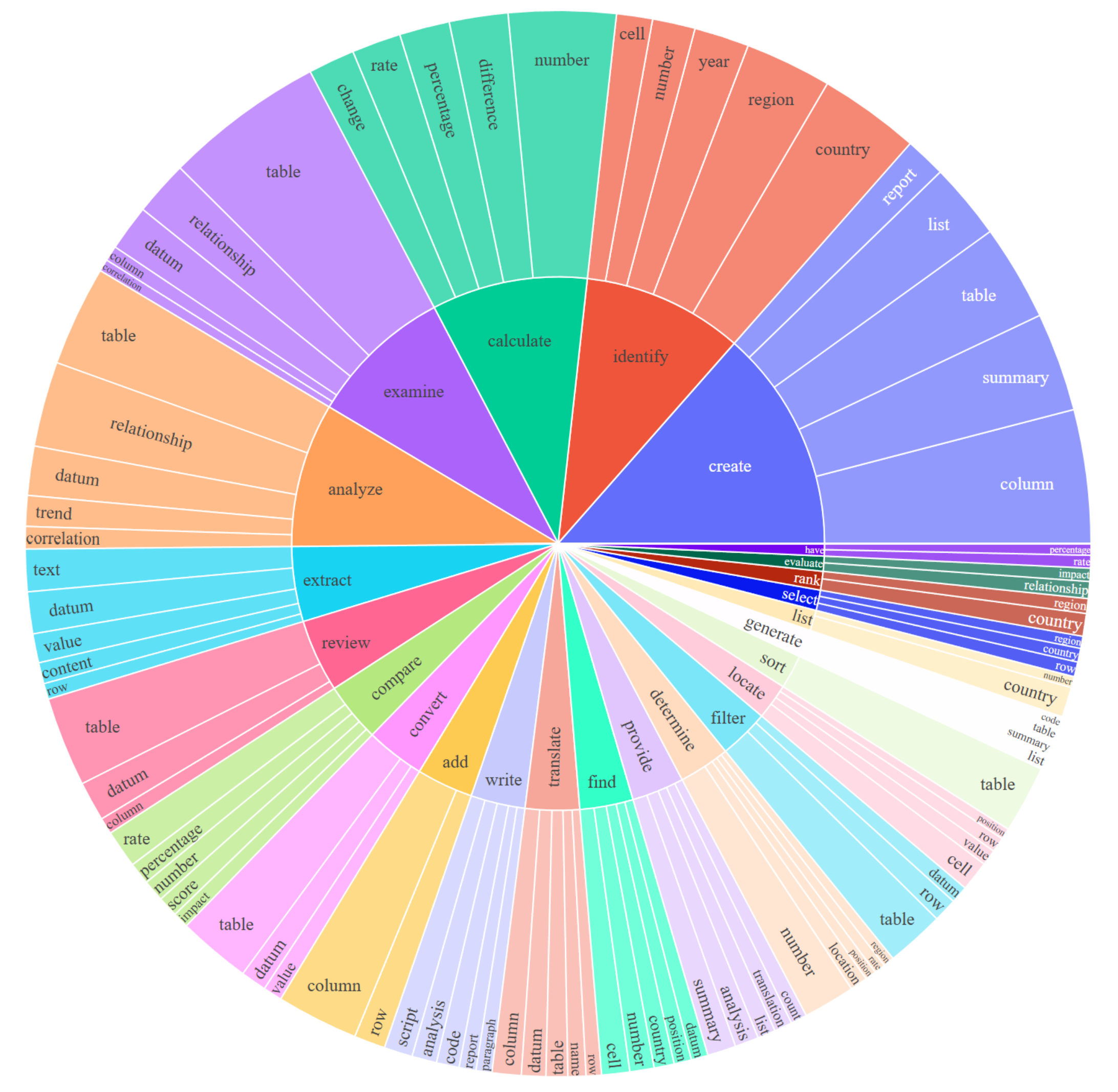}
  \caption{The top 25 most prevalent root verbs (the inner circle) and their top 5 direct nouns (the outer circle) in the synthetic instructions of TableDreamer-27K.}
  \label{inst_root_verb}
\end{figure}

\subsection{Table Generation}

Existing general data synthesis methods like self-instruct can not fully capture the complexity and diversity inherent in structured tabular data, leading to limited variety of synthesized tables. Therefore, we meticulously design a table synthesis prompt that fully considers the important table attributes. First of all, various \textbf{topics}, \textbf{subtopics} and corresponding \textbf{table titles} of different domains are elicited from an established LLM, which then serve as the guidance for generating table content of different domains. For example, given the topic `Science and Technology' and the subtopic `AI Applications', a viable table title could be `Detailed Analysis of AI Integration in Auto. Vehicles, 2022'.

On this basis, we further incorporate key table attributes in the prompt to enhance the diversity of synthetic tables. (1) \textbf{table type}. We randomly sample one table type from three common candidates including flat tables, horizontal tables and hierarchical tables~\cite{HiTab,liu2024rethinking_tu_llm,infotabs}. (2) \textbf{table size}. We randomly choose the row number and the column number of the table within an appropriate range to create tables of various sizes. (3) \textbf{header structure}. For hierarchical tables with multi-level row headers and column headers, we randomly appoint the expected row header and column header structure from common combinations. For instance, a hierarchical table could have a 3-level column header and 2-level row header. (4) \textbf{cell relation}. There may be dependency relations between different table cells, e.g., in a business revenue table, the value of `net profit' should be the difference between the `revenue' and the `cost'. Thus, we require the LLM to utilize markdown formulas to represent such relations in the target cells if necessary, which can be automatically extracted and computed by scripts to obtain the final results. (5) \textbf{table format}. We use the HTML format to represent the synthesized hierarchical tables in order to accurately reflect merged cells and hierarchical headers and the Markdown-style format to represent flat and horizontal tables.

Taking into account the above table attributes, we employ the LLM as a table generator to synthesize diverse tables, which are further processed to compute results of potential formulas and are filtered to remove invalid tables such as incomplete tables with missing cells.

\subsection{Instruction Tuning Data Generation}
To provide a better foundation for instruction generation, we collect 20 different table understanding tasks and their descriptions from published studies~\cite{tu_paradigm_survey,sui2024table_meets_llm,zhao-etal-2022-reastap,zhao-etal-2023-investigating_table_to_text}, such as table-based numerical reasoning, table structure understanding and so on. The full list of seed tabular tasks are shown in the Table \ref{seed_task_descriptions}. On the basis of synthetic tables and the task descriptions, we use the LLM to generate a set of task instructions which serve as the initial seed instructions for subsequent data evolution.

\paragraph{Input Space Exploration.} To achieve a more comprehensive exploration of the input space, each sample in the seed data will undergo LLM-based data evolution in three directions respectively, thereby synthesizing more diverse data.

\textbf{Instruction Complication.} Inspired by previous instruction generation methods~\cite{xu2023wizardlm,mmevol}, we devise different evolution strategies to create more complex instructions based on the original table and the instruction. For instance, `increasing the task number' will create new instructions that ask the LLM to complete multiple tabular tasks at once, and `adding the reasoning steps' will generate multi-step problems. As LLMs' capabilities continue to improve, increasing the difficulty of input instructions assists us in uncovering the potential weaknesses in the table understanding ability of state-of-the-art LLMs, which enables us to enhance the model's capabilities in a more targeted manner.

\textbf{Instruction Generalization.} Considering that the instructions in the seed data are primarily limited to 20 predefined tabular tasks, we use the LLM to synthesize instructions of new tasks that are different from the original ones. We find that the LLM could create instructions of interesting and creative tabular tasks, e.g., analyzing the original table and providing recommendations, translating several columns into a new language and so on. Such task instructions are often not included in the public table-related datasets but can greatly improve the diversity of the instruction tuning data. In addition to generating new tabular task instructions, we also generate instructions that possess the same task type to the original one in order to improve model robustness towards instruction variations.

\textbf{Table Generalization.} Prior studies have found that current LLMs lack the robustness towards content and structural perturbations of input tables~\cite{liu2024rethinking_tu_llm,zhou-etal-2024-freb_tqa,robustness_of_tsu}. For instance, LLMs may experience significant performance fluctuations with changes in table formats and the order of rows and columns. This robustness is crucial for the practical application of tabular LLMs, as input tables from real-world users can vary greatly. To this end, we design table evolution strategies to create more table variations based on previously synthesized tables, e.g., changing the original table format, modifying the table header, reordering rows and columns and so on. This table generalization further improves the table diversity in the final training data which helps the model learn to maintain its performance despite these perturbations.

\paragraph{Weakness Data Identification.} Although the input space exploration can generate a large variety of data, some of these samples may already be well-handled by the target LLM. Fine-tuning with such data could yield little performance improvements while consuming additional training resources. Thus, we utilize the LLM-as-a-judge~\cite{llm_as_a_judge} to evaluate the response from the target LLM and identify samples where the target LLM underperforms. Concretely, given the response from the target LLM (e.g., Llama3.1-8B-instruct) and the reference response from a more powerful LLM (e.g., GPT-4o), an LLM rates the correctness of the model response on a 5-point likert scale, with lower scores indicating poorer performance. The samples with scores below 3 points are considered as weakness data, which will be used as the seed data for the next round of input space exploration and thus guide the overall data synthesis direction towards valuable data points that expose the model's deficiencies in table understanding ability. This iterative process between the input space exploration and the weakness detection can be performed multiple times, and the accumulated weakness data together with reference responses are used as the final table instruction tuning data.

\subsection{Dataset Statistics and Cases}
Unless otherwise specified, we use GPT-4o to synthesize tables, instructions and corresponding responses and select the Llama3.1-8B-instruct as the target LLM for weakness data detection.  Starting from 3,272 seed data over 1,541 synthetic tables, we perform 2 rounds of iterative data synthesis process, ending in 27,083 instruction tuning data over 7,950 tables after filtering the invalid samples (e.g., failed data evolution results), which is denoted as TableDreamer-27K. Besides, we also replace GPT-4o with Llama3.1-70B-instruct to synthesize 27K training data, which is used for a fair comparison with other data synthesis baselines that we also reimplemented with Llama3.1-70B-instruct. Figure \ref{dataset_examples_1} demonstrates an example of the synthetic data. The diversity of the generated 27K instructions from GPT-4o is illustrated in Figure \ref{inst_root_verb}, where we plot the top 25 most prevalent root verbs and their top 5 direct
nouns that appear at least 15 times. We can find that TableDreamer could generate diverse instructions and tables that encompass a broad range of tabular tasks and domains. More statistics, examples and comparison between different synthetic table instruction tuning datasets are given in App. \ref{sec:more_dataset_statistics}. The detailed data evolution strategies and prompts are shown in App. \ref{TableDreamer_implementations}.

\section{Experiments}
\label{sec:experiments}
\subsection{Experimental Setup}
\textbf{Benchmarks.} We select 9 public benchmarks: TABMWP~\cite{tabmwp}, WTQ~\cite{WTQ}, HiTab~\cite{HiTab}, AIT-QA~\cite{aitqa}, TabMCQ~\cite{tabmcq}, TabFact~\cite{TabFact}, InfoTabs~\cite{infotabs}, FeTaQA~\cite{FeTaQA} and QTSumm~\cite{qtsumm}, which cover three tasks including table question answering (TQA), table-based fact verification (TFV) and table-to-text generation (T2T). The original question and the table in these benchmarks are serialized into an input text with various instruction templates and four common table formats (HTML, Markdown, csv, tsv) for evaluating the LLM's robust table understanding ability. Besides, we also consider the synthetic benchmark from TableGPT~\cite{tablegpt_microsoft} which contains many unusual tabular tasks such as data imputation and thus can be used to evaluate the model’s out-of-distribution (OOD) generalization ability. All selected benchmarks are shown in Table \ref{benchmark_descriptions}.

\begin{table*}[t]\footnotesize
\centering
\renewcommand{\arraystretch}{1.3}
\setlength\tabcolsep{2pt}
\scalebox{0.7}{
\begin{tabular}{c|c|ccccc|cc|cc|c|c} 
\hline
\multirow{2}{*}{\textbf{Method}} & \multirow{2}{*}{\textbf{\# IFT Data}} & \multicolumn{5}{c|}{\textbf{TQA}} & \multicolumn{2}{c|}{\textbf{TFV}} & \multicolumn{2}{c|}{\textbf{T2T}} & \multirow{2}{*}{\textbf{TableGPT}} & \multirow{2}{*}{\begin{tabular}[c]{@{}c@{}}\textbf{Ave.}\\\textbf{Acc.}\end{tabular}} \\ 
\cline{3-11}
 &  & \textbf{TABMWP} & \textbf{WTQ} & \textbf{HiTab} & \textbf{AIT-QA} & \textbf{TabMCQ} & \textbf{TabFact} & \textbf{InfoTabs} & \textbf{FeTaQA} & \textbf{QTSumm} &  &  \\ 
\hline
\multicolumn{13}{l}{{\cellcolor[rgb]{0.957,0.957,0.957}}\textit{LLM}} \\
Baichuan2-7B-Chat & - & 30.31 & 4.60 & 1.58 & 10.95 & 41.59 & 14.39 & 19.83 & 57.86 & 30.24 & 18.14 & 22.95 \\
GLM4-9B-Chat & - & 39.87 & 20.30 & 8.94 & 36.98 & 43.57 & 11.99 & 11.16 & 77.73 & 55.19 & 42.53 & 34.83 \\
DeepSeek-V2-Lite-16B-Chat & - & 49.01 & 15.65 & 7.67 & 29.94 & 63.45 & 29.75 & 37.11 & 64.20 & 35.81 & 29.36 & 36.19 \\
Phi3.5-mini-3.8B & - & 59.45 & 19.26 & 7.99 & 35.02 & 64.72 & 35.60 & 43.37 & 77.75 & 57.14 & 8.05 & 40.83 \\
MiniCPM3-4B & - & 50.53 & 34.06 & 20.93 & 55.34 & 72.98 & 28.09 & 42.33 & 68.55 & 42.39 & 40.79 & 45.60 \\
Mistral-7B-Instruct-v0.3 & - & 37.92 & 25.71 & 16.41 & 52.05 & 57.82 & 47.80 & 42.68 & 78.63 & 55.57 & 44.26 & 45.88 \\
InternLM2.5-7B-Chat & - & 50.22 & 32.59 & 13.51 & 51.46 & 36.25 & 45.07 & 47.33 & 81.43 & 62.86 & 39.52 & 46.02 \\
Yi-1.5-9B-Chat & - & 31.45 & 38.23 & 14.02 & 51.85 & 55.97 & 46.15 & 46.22 & 82.03 & 59.18 & 42.15 & 46.73 \\
Llama3.1-8B-Instruct & - & 53.39 & 36.53 & 11.35 & 43.63 & 75.31 & 53.87 & 48.94 & 78.98 & 66.98 & 21.68 & 49.07 \\
\multicolumn{13}{l}{{\cellcolor[rgb]{0.957,0.957,0.957}}\textit{General Instruction Tuning Data Synthesis Methods}} \\
Self-Instruct & 100K & 46.68 & 28.98 & 13.77 & 48.92 & 80.27 & 52.92 & 45.07 & 81.13 & 53.48 & 43.13 & 49.44 \\
Dynasour & 132K & 49.71 & 28.59 & 20.11 & 43.44 & 59.66 & 50.70 & 41.01 & 57.56 & 42.57 & 13.40 & 40.67 \\
GenQA & 100K & 59.87 & 41.06 & \uline{21.63} & \textbf{57.14} & 70.35 & 55.01 & 39.38 & 67.05 & 56.49 & 32.94 & 50.09 \\
Evol-Instruct & 100K & 54.61 & 31.83 & 12.37 & 45.20 & 73.27 & 54.12 & 45.61 & 83.02 & 62.77 & 42.55 & 50.54 \\
Magpie & 100K & 57.11 & 34.66 & 13.89 & 47.16 & 76.96 & 51.21 & 43.83 & 80.02 & \textbf{76.90} & 40.59 & 52.23 \\
\multicolumn{13}{l}{{\cellcolor[rgb]{0.957,0.957,0.957}}\textit{Table Instruction Tuning Data Synthesis Methods}} \\
OmniTab & 100K & 17.53 & 22.67 & 18.84 & 35.02 & 50.63 & 16.37 & 3.14 & 5.04 & 4.82 & 18.38 & 19.24 \\
ReasTap & 100K & 11.22 & 19.54 & 9.96 & 20.54 & 48.49 & 15.66 & 5.70 & 7.14 & 4.92 & 20.67 & 16.38 \\
UCTR & 43K & 17.61 & 12.03 & 8.84 & 17.31 & 35.76 & 20.96 & 20.35 & 15.23 & 7.51 & 7.09 & 16.27 \\
TableGPT-syn-data & 66K & 25.21 & 16.13 & 9.13 & 24.26 & 47.52 & 19.70 & 25.29 & 46.03 & 36.64 & \textbf{47.23}$^\dagger$ & 29.71 \\
TableLLM-syn-data & 80K & 46.10 & 42.24$^\dagger$ & 13.92 & 39.72 & 25.46 & 29.24 & 31.31 & 79.08$^\dagger$ & 55.94 & 23.74 & 38.68 \\
\multicolumn{13}{l}{{\cellcolor[rgb]{0.957,0.957,0.957}}\textit{Tabular LLM}} \\
TableBenchLLM (Llama3.1-8B) & 20K & 25.83 & 18.50$^\dagger$ & 12.31 & 29.74$^\dagger$ & 30.41 & 23.97$^\dagger$ & 17.33 & 48.27$^\dagger$ & 42.30 & 16.78 & 26.54 \\
TableLLM (CodeLlama-7B) & 80K & 43.11 & 37.86$^\dagger$ & 15.67 & 45.40 & 24.87 & 30.47 & 27.55 & 67.35$^\dagger$ & 37.66 & 15.14 & 34.51 \\
\textcolor[rgb]{0.502,0.502,0.502}{TableGPT2 (Qwen2.5-7B)}$^\ddagger$ & \textcolor[rgb]{0.502,0.502,0.502}{2.36M} & \textcolor[rgb]{0.502,0.502,0.502}{56.35} & \textcolor[rgb]{0.502,0.502,0.502}{49.35} & \textcolor[rgb]{0.502,0.502,0.502}{38.26} & \textcolor[rgb]{0.502,0.502,0.502}{73.97} & \textcolor[rgb]{0.502,0.502,0.502}{85.71} & \textcolor[rgb]{0.502,0.502,0.502}{60.42} & \textcolor[rgb]{0.502,0.502,0.502}{54.87} & \textcolor[rgb]{0.502,0.502,0.502}{84.72} & \textcolor[rgb]{0.502,0.502,0.502}{64.10} & \textcolor[rgb]{0.502,0.502,0.502}{70.25} & \textcolor[rgb]{0.502,0.502,0.502}{63.80} \\
\multicolumn{13}{l}{{\cellcolor[rgb]{0.957,0.957,0.957}}\textit{Ours}} \\
\textbf{TableDreamer} (Llama3.1-70B-Instruct) & 27K & \uline{60.57} & \uline{42.47} & 17.25 & \uline{56.75} & \uline{82.99} & \uline{57.32} & \uline{49.98} & \textbf{84.67} & 75.12 & 33.03 & \uline{56.02} \\
\textbf{TableDreamer} (GPT-4o) & 27K & \textbf{64.61} & \textbf{54.66} & \textbf{22.88} & 53.22 & \textbf{84.29} & \textbf{63.09} & \textbf{57.65} & \uline{84.37} & \uline{75.97} & \uline{46.20} & \textbf{60.69} \\
\hline
\end{tabular}
}
\caption{Evaluation results on 10 tabular task benchmarks. $\dagger$ indicates that the model's fine-tuning data includes training samples from the corresponding dataset. $\ddagger$: we only list the performance of the TableGPT2 as its training data already contains these common benchmark datasets and the data volume also far exceeds others. } 
\label{main_results}
\end{table*}

\textbf{Evaluation Metrics.} For TQA, TFV and TableGPT benchmarks, the input instructions ask LLMs to output the final answer in the JSON format, which can be automatically extracted with regular expressions to compute exact match accuracy. For T2T benchmarks that are hard to accurately evaluate the correctness of the model response with automatic text generation metrics like BLEU~\cite{bleu_metric}, we use LLM-as-a-judge evaluation, where GPT-4o-mini determines the accuracy of the model’s responses based on the gold answer. The zero-shot setting is adopted for 9 public benchmarks except the TableGPT, as it provides test data in zero-shot and few-shot settings. Thus we report the average accuracy of two settings. 

\textbf{Baselines.} We consider baselines of four genres. \textbf{(1) General LLMs} such as Llama3.1-8B-instruct~\cite{llama3_herdmodels} and Mistral-7B-Instruct-v0.3~\cite{jiang2023mistral7b}. \textbf{(2) General Instruction Tuning Data Synthesis Methods} including the Self-Instruct~\cite{wang-etal-2023-self-instruct}, Dynasour~\cite{yin-etal-2023-dynosaur}, Evol-Instruct~\cite{xu2023wizardlm}, GenQA~\cite{GenQA} and Magpie~\cite{magpie}. \textbf{(3) Data Synthesis Methods for Table Instruction Tuning}. We consider traditional tabular question generation methods including the OmniTab~\cite{jiang-etal-2022-omnitab}, ReasTap~\cite{zhao-etal-2022-reastap} and UCTR-ST~\cite{UCTR-ST}, as well as recent LLM-based synthetic data from the TableGPT~\cite{tablegpt_microsoft} and the TableLLM~\cite{tablellm}, which use GPT-3.5 to generate instructions based on public tables. \textbf{(4) Tabular LLMs} including the TableBenchLLM~\cite{tablebenchllm} which is fine-tuned from Llama3.1-8B-base with 20K manually collected data, and the TableLLM~\cite{tablellm} which is fine-tuned from CodeLlama-7B with 80K synthetic data. We also evaluate the powerful TableGPT2-7B~\cite{tablegpt2_zhejiang} that is fine-tuned from Qwen2.5-7B-instruct~\cite{qwen2_5_technical_report} with 2.36M in-house query-table-output tuples. Implementation details are given in the Appendix \ref{baseline_implementations}.

\begin{table*}[t]\footnotesize
\centering
\renewcommand{\arraystretch}{1.3}
\setlength\tabcolsep{2pt}
\scalebox{0.75}{
\begin{tabular}{c|ccc|cc|cc|cc|c} 
\hline
\multirow{2}{*}{\begin{tabular}[c]{@{}c@{}}\textbf{\# Available Train Data }\\\textbf{of Each Dataset}\end{tabular}} & \multicolumn{3}{c|}{\textbf{TQA}} & \multicolumn{2}{c|}{\textbf{TFV}} & \multicolumn{2}{c|}{\textbf{T2T}} & \multicolumn{2}{c|}{\textbf{Held-out}} & \multirow{2}{*}{\textbf{Ave. Acc}} \\ 
\cline{2-10}
 & \textbf{TABMWP} & \textbf{WTQ} & \textbf{HiTab} & \textbf{TabFact} & \textbf{InfoTabs} & \textbf{FeTaQA} & \textbf{QTSumm} & \textbf{AIT-QA} & \textbf{TabMCQ} &  \\ 
\hline
Llama3.1-8B-Instruct & 53.39 & 36.53 & 11.35 & 53.87 & 48.94 & 78.98 & 66.98 & 43.63 & 75.31 & 50.01 \\ 
\hline
20 & 55.91 & 37.43 & 12.81 & 56.50 & 47.62 & 84.57 & 72.24 & 46.77 & 76.96 & 52.44 \\
w/ TableDreamer-27K & 64.88 & 56.23 & 24.17 & 60.46 & 53.38 & \textbf{83.87} & \textbf{76.62} & 53.42 & 83.28 & 59.94 \\
$\triangle$ & {\cellcolor[rgb]{0.792,1,0.792}}8.97 & {\cellcolor[rgb]{0.675,1,0.675}}18.80 & {\cellcolor[rgb]{0.675,1,0.675}}11.36 & {\cellcolor[rgb]{0.824,0.949,0.824}}3.96 & {\cellcolor[rgb]{0.792,1,0.792}}5.76 & {\cellcolor[rgb]{1,0.835,0.835}}-0.70 & {\cellcolor[rgb]{0.824,0.949,0.824}}4.38 & {\cellcolor[rgb]{0.792,1,0.792}}6.65 & {\cellcolor[rgb]{0.792,1,0.792}}6.32 & {\cellcolor[rgb]{0.792,1,0.792}}7.50 \\ 
\hline
50 & 56.18 & 37.75 & 14.78 & 56.34 & 47.88 & 83.23 & 69.48 & 51.07 & 77.84 & 52.23 \\
w/ TableDreamer-27K & 70.89 & \textbf{56.37} & 26.90 & 60.68 & 47.22 & 83.37 & 74.95 & 61.64 & 83.86 & 60.05 \\
$\triangle$ & {\cellcolor[rgb]{0.675,1,0.675}}14.71 & {\cellcolor[rgb]{0.675,1,0.675}}18.62 & {\cellcolor[rgb]{0.675,1,0.675}}12.12 & {\cellcolor[rgb]{0.824,0.949,0.824}}4.34 & {\cellcolor[rgb]{1,0.835,0.835}}-0.66 & {\cellcolor[rgb]{0.937,1,0.937}}0.14 & {\cellcolor[rgb]{0.792,1,0.792}}5.47 & {\cellcolor[rgb]{0.675,1,0.675}}10.57 & {\cellcolor[rgb]{0.792,1,0.792}}6.02 & {\cellcolor[rgb]{0.792,1,0.792}}7.82 \\ 
\hline
100 & 56.77 & 40.69 & 23.28 & 48.04 & 45.25 & 77.57 & 55.43 & 55.77 & 68.12 & 49.58 \\
w/ TableDreamer-27K & 70.96 & 54.37 & 36.04 & 57.07 & 46.00 & 81.38 & 73.28 & \textbf{64.18} & 84.15 & 59.87 \\
$\triangle$ & {\cellcolor[rgb]{0.675,1,0.675}}14.19 & {\cellcolor[rgb]{0.675,1,0.675}}13.68 & {\cellcolor[rgb]{0.675,1,0.675}}12.76 & {\cellcolor[rgb]{0.792,1,0.792}}9.03 & {\cellcolor[rgb]{0.937,1,0.937}}0.75 & {\cellcolor[rgb]{0.824,0.949,0.824}}3.81 & {\cellcolor[rgb]{0.675,1,0.675}}17.85 & {\cellcolor[rgb]{0.792,1,0.792}}8.41 & {\cellcolor[rgb]{0.675,1,0.675}}16.03 & {\cellcolor[rgb]{0.675,1,0.675}}10.30 \\ 
\hline
200 & 66.43 & 40.01 & 32.61 & 61.66 & 52.29 & 71.34 & 40.82 & 57.72 & 76.48 & 52.17 \\
w/ TableDreamer-27K & \textbf{76.59} & 50.59 & \textbf{41.94} & \textbf{63.33} & \textbf{57.44} & 78.43 & 72.26 & 59.29 & \textbf{84.64} & \textbf{62.94} \\
$\triangle$ & {\cellcolor[rgb]{0.675,1,0.675}}10.16 & {\cellcolor[rgb]{0.675,1,0.675}}10.58 & {\cellcolor[rgb]{0.792,1,0.792}}9.33 & {\cellcolor[rgb]{0.824,0.949,0.824}}1.67 & {\cellcolor[rgb]{0.792,1,0.792}}5.15 & {\cellcolor[rgb]{0.792,1,0.792}}7.09 & {\cellcolor[rgb]{0.49,1,0.49}}31.44 & {\cellcolor[rgb]{0.824,0.949,0.824}}1.57 & {\cellcolor[rgb]{0.792,1,0.792}}8.16 & {\cellcolor[rgb]{0.675,1,0.675}}10.77 \\ 
\hline
\end{tabular}
}
\caption{Evaluation results under the few-shot learning setting, where only a limited number of training samples from 7 datasets (the first 7 columns) are available and TableDreamer data is used as additional training data.} 
\label{low_resource_results}
\end{table*}

\begin{table}[t]\footnotesize
\centering
\renewcommand{\arraystretch}{1.3}
\setlength\tabcolsep{2pt}
\scalebox{0.7}{
\begin{tabular}{c|c|cccc|c} 
\hline
\textbf{Mehtod} & \textbf{\# IFT Data} & \textbf{TQA} & \textbf{TFV} & \textbf{T2T} & \textbf{TableGPT} & \textbf{Ave. Acc} \\ 
\hline
Llama3.1-8B-Instruct & - & 44.04 & 51.41 & 72.98 & 21.68 & 49.07 \\ 
\hline
w/ TableDreamer & 27K & \textbf{55.93} & \textbf{60.37} & \textbf{80.17} & 46.20 & \textbf{60.69} \\ 
\hline
w/o Flat Tables & \multirow{2}{*}{17K} & 51.41 & 52.02 & 74.85 & 40.59 & 55.13 \\
$\triangle$ &  & {\cellcolor[rgb]{1,0.675,0.675}}-4.53 & {\cellcolor[rgb]{1,0.592,0.592}}-8.36 & {\cellcolor[rgb]{1,0.675,0.675}}-5.33 & {\cellcolor[rgb]{1,0.675,0.675}}-5.61 & {\cellcolor[rgb]{1,0.675,0.675}}-5.56 \\ 
\hline
w/o Hier. Tables & \multirow{2}{*}{17K} & 49.24 & 52.54 & 76.37 & \textbf{46.79} & 55.08 \\
$\triangle$ &  & {\cellcolor[rgb]{1,0.675,0.675}}-6.69 & {\cellcolor[rgb]{1,0.675,0.675}}-7.83 & {\cellcolor[rgb]{1,0.675,0.675}}-3.80 & {\cellcolor[rgb]{0.835,1,0.835}}+0.59 & {\cellcolor[rgb]{1,0.675,0.675}}-5.61 \\ 
\hline
w/o Hori. Tables & \multirow{2}{*}{18K} & 54.58 & 51.40 & 78.07 & 45.38 & 57.72 \\
$\triangle$ &  & {\cellcolor[rgb]{1,0.792,0.792}}-1.35 & {\cellcolor[rgb]{1,0.592,0.592}}-8.98 & {\cellcolor[rgb]{1,0.792,0.792}}-2.11 & {\cellcolor[rgb]{1,0.792,0.792}}-0.82 & {\cellcolor[rgb]{1,0.792,0.792}}-2.97 \\ 
\hline
w/o Data Evolution & \multirow{2}{*}{3K} & 47.71 & 49.50 & 71.28 & 38.68 & 51.88 \\
$\triangle$ &  & {\cellcolor[rgb]{1,0.592,0.592}}-8.22 & {\cellcolor[rgb]{1,0.592,0.592}}-10.87 & {\cellcolor[rgb]{1,0.592,0.592}}-8.90 & {\cellcolor[rgb]{1,0.675,0.675}}-7.52 & {\cellcolor[rgb]{1,0.592,0.592}}-8.82 \\ 
\hline
~w/o Inst. Gene. & \multirow{2}{*}{18K} & 52.32 & 51.77 & 78.26 & 40.89 & 56.26 \\
$\triangle$ &  & {\cellcolor[rgb]{1,0.675,0.675}}-3.61 & {\cellcolor[rgb]{1,0.592,0.592}}-8.60 & {\cellcolor[rgb]{1,0.792,0.792}}-1.91 & {\cellcolor[rgb]{1,0.792,0.792}}-5.31 & {\cellcolor[rgb]{1,0.675,0.675}}-4.44 \\ 
\hline
~w/o Inst. Comp. & \multirow{2}{*}{18K} & 50.83 & 51.25 & 73.95 & 39.82 & 54.44 \\
$\triangle$ &  & {\cellcolor[rgb]{1,0.675,0.675}}-5.10 & {\cellcolor[rgb]{1,0.592,0.592}}-9.12 & {\cellcolor[rgb]{1,0.675,0.675}}-6.22 & {\cellcolor[rgb]{1,0.675,0.675}}-6.38 & {\cellcolor[rgb]{1,0.675,0.675}}-6.26 \\ 
\hline
~w/o Table Gene. & \multirow{2}{*}{19K} & 50.20 & 54.29 & 76.19 & 42.35 & 55.43 \\
$\triangle$ &  & {\cellcolor[rgb]{1,0.675,0.675}}-5.73 & {\cellcolor[rgb]{1,0.675,0.675}}-6.09 & {\cellcolor[rgb]{1,0.675,0.675}}-3.98 & {\cellcolor[rgb]{1,0.675,0.675}}-3.85 & {\cellcolor[rgb]{1,0.675,0.675}}-5.26 \\ 
\hline
w/o Weakness Iden. & \multirow{2}{*}{34K} & 53.12 & 51.72 & 75.82 & 42.12 & 56.28 \\
$\triangle$ &  & {\cellcolor[rgb]{1,0.792,0.792}}-2.81 & {\cellcolor[rgb]{1,0.592,0.592}}-8.65 & {\cellcolor[rgb]{1,0.675,0.675}}-4.35 & {\cellcolor[rgb]{1,0.675,0.675}}-4.08 & {\cellcolor[rgb]{1,0.675,0.675}}-4.41 \\
\hline
\end{tabular}

}
\caption{Ablation experiment results. We report average accuracy on four task types. $\triangle$ stands for the performance gap between the Llama3.1-8B-Instruct finetuned with TableDreamer data and its variants. `Hier.' and `Hori.' stands for hierarchical and horizontal tables. `Inst. Gene.', `Inst. Comp.', `Table. Gene.' and `Weakness Iden.' represents three data evoluation directions and weakness data identification respectively.} 
\label{ablation_study_results}
\end{table}

\subsection{Results and Analysis}
\paragraph{Main Results.} \textit{Performance of general LLMs.} As shown in Table \ref{main_results}, recent LLMs demonstrate varying proficiency in the table understanding ability, with the Llama3.1-8B-instruct exhibiting the best performance while models like Baichuan2-7B-Chat showing comparatively weaker performance. Their performance difference is likely due to the construction table-related fine-tuning data during the post-training stage. Moreover, we can find that small language model can also possess great table understanding ability, e.g., MiniCPM3-4B achieves better performance than large models like GLM4-9B-Chat, which opens up new possibilities for developing powerful and efficient tabular LLMs.

\textit{Performance of tabular LLMs.} Compared with general LLMs, recent tabular LLMs such as TableBenchLLM exhibit surprisingly poorer performance on the benchmarks where they should be experts, even after being fine-tuned with the corresponding training dataset. Moreover, they can not effectively handle the unseen tabular tasks in the TableGPT benchmark. This shows that these tabular LLMs actually possess limited generalization ability especially out-of-distribution generalization, which is consistent with the findings from~\citet{rethinking_table_instruction_tuning}. After a careful inspection, we find that this is due to the insufficient diversity in their instruction tuning data, e.g., the training data of TableBenchLLM only contain flat tables with a fixed Python dictionary-style table format and the instructions are primally limited to pre-defined tabular tasks. As a result, they can only perform well under the in-distribution setting, which highly constrains their application scenarios. By contrast, the TableGPT2 delivers the best overall results particularly on the TableGPT benchmark, showcasing the effectiveness of the 2.36M in-house high-quality training data, which includes not only public tabular datasets but also substantial synthetic data that are further refined by human annotators. 

\textit{Performance of data synthesis methods.} General instruction tuning data synthesis methods could be successfully extended to generate table instruction tuning data and bring considerable performance boost. For instance, fine-tuning with 100K Magpie synthetic data boosts the average accuracy from 49.07\% to 52.23\%. The traditional question generation approaches such as ReasTap obtain the worst performance because they can only generate simple table-related questions either through predefined question templates or by converting SQL queries. In comparison, although the LLM-based synthetic data from TableGPT and TableLLM can enhance the in-distribution model performance, e.g., fine-tuning with TableGPT synthetic data achieves the best result on the corresponding TableGPT benchmark, they still fail to improve the out-of-distribution table understanding capability on other benchmarks, which eventually yield a degenerated overall performance.

\textit{Effectiveness of TableDreamer.} With Llama3.1-70B-instruct as the data synthesis LLM, TableDreamer improves the average accuracy of Llama3.1-8B-instruct by 6.95\% ($49.07\%\rightarrow56.02\%$) ands surpasses other baselines without using any data from the public benchmarks, which validates the effectiveness of the proposed framework. The performance boost increases to 11.62\% with the GPT-4o synthetic data due to better data quality. Notably, TableDreamer achieves a strong result (46.20\%) on the TableGPT benchmark and is comparable to the model fine-tuned with TableGPT training data (47.23\%), which showcases its effectiveness in improving the out-of-distribution table understanding capability. Moreover, TableDreamer obtains superior results with better data efficiency than data synthesis baselines, and is even competitive with the
powerful TableGPT2 fine-tuned with 2.36M high-quality data.

\paragraph{TableDreamer as Data Augmentation.} As shown in the Table \ref{low_resource_results}, fine-tuning the model with very little labeled data offers limited improvement compared with the original performance, and adding TableDreamer synthetic data can bring a significant performance boost across various few-shot learning settings, which demonstrates its effectiveness in mitigating the scarcity of annotated table instruction tuning data.

\begin{figure*}[t]
  \centering
  \includegraphics[width=\linewidth]{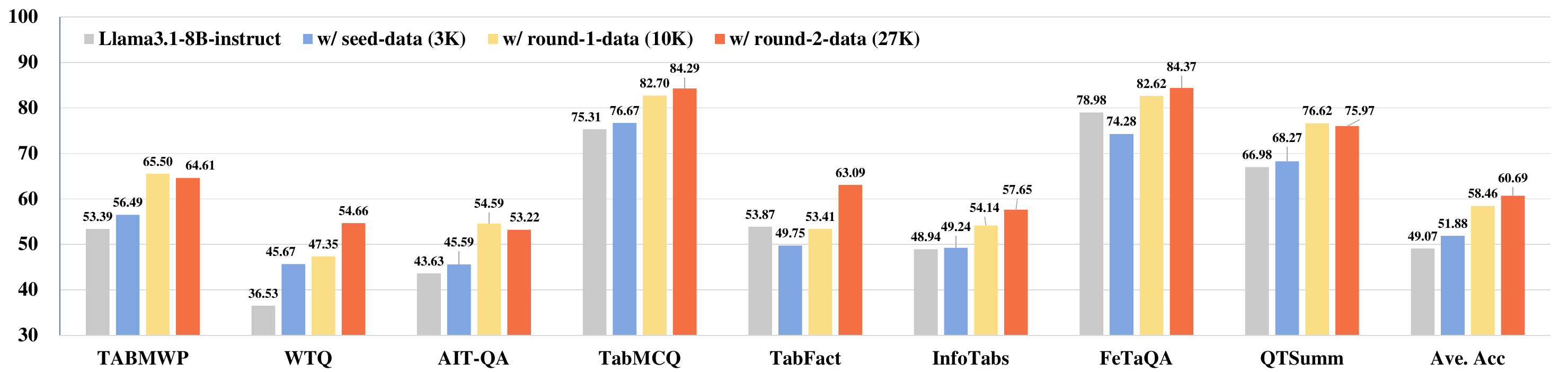}
  \caption{The performance improvement as the TableDreamer synthetic data (from GPT-4o) continues to accumulate.}
  \label{data_evol_influence}
\end{figure*}

\paragraph{Ablation Study.} (1) \textit{Ablation of synthetic tables.}  We remove one type of tables and related instruction tuning data from the total data to analyze their influence, respectively. As presented in Table \ref{ablation_study_results}, removing each type of synthetic tables will cause negative effects due to the degenerated table diversity. We also observe the similar phenomenon in the main experiments where the fine-tuning with TableGPT-syn-data (only including flat tables) results in poor performance on tables of different types (e.g., hierarchical tables from HiTab). Compared with others, removing horizontal tables leads to a lower performance decrease which may be because most benchmarks only contain flat or hierarchical tables.  (2) \textit{Ablation of data evolution.} We remove the data generated from different data evolution directions. We can find that all three data evolution directions make substantial contributions to the final model performance, and `w/o Instruction Complication' causes a more significant performance decline than others, which highlights the importance of complex instructions in enhancing the model's table understanding ability. Unsurprisingly, `w/o Data Evolution' causes the worst performance as we only fine-tuned the model with 3K seed data. This shows that, simply using LLMs to synthesize instructions of pre-defined types, which is the common practice of recent tabular LLMs, is insufficient to improve the model performance, and we need to thoroughly explore the vase input space for better data diversity. (3) \textit{Ablation of weakness data identification.} We use all generated data from data evolution for fine-tuning rather than the selected weakness data. Despite using more synthetic data (34K), the model actually suffers a performance drop of 4.41, which suggests that choosing the weakness-exposing data is more conducive to model performance than blindly increasing the data volume.

\paragraph{Effect of Data Size.} Since TableDreamer synthesizes instruction tuning data through an iterative collaboration between input space exploration and weakness data identification, we investigate the performance improvement resulting from the accumulation of synthetic data. To this end, we fine-tuned the Llama3.1-8B-Instruct with the initial seed data (3K), the accumulated synthetic data after the first round (10K, including seed data) and the total data after the second round (27K), respectively. From the results in Figure \ref{data_evol_influence}, we can observe that the average performance continues to improve with the growth of synthetic data, which demonstrates the scalability of the method.

\paragraph{Performance of Recent R1-style Reasoning Models}
The recent emerging reasoning LLMs (o1 and R1-style) have achieved significant progress on complicated math, code, and other tasks that demand human-level complex reasoning ability. However, their capability to understand structured tabular data has not been thoroughly investigated. To fill the gap, we evaluate representative reasoning LLMs on 9 tabular benchmarks except TableGPT benchmark to save API cost of R1 and GPT-4o. 

From the results shown in Table \ref{reasoning_LLM_results}, we can find that reasoning LLMs (like DeepSeek-R1 and QwQ-32B) surpass traditional LLMs (like DeepSeek-V3 and GPT-4o) and achieve the best overall performance, which demonstrates that improving general reasoning ability of LLMs can also boost their table understanding skills, e.g., DeepSeek-R1 improves the average accuracy of DeepSeek-V3 by 3.20 (71.25$\rightarrow$74.45). The QwQ-32B even performs slightly better than 671B DeepSeek-R1, which could be attributed to the reason that the Qwen-2.5 backbone has been specially enhanced for understanding table data~\cite{qwen2_5_technical_report}.

The R1-distilled smaller LLMs, which were fine-tuned with R1's 800K SFT data, also outperform their vanilla versions, e.g., R1-Distill-Llama-8B beats Llama3.1-8B-instruct by 7.12 in average accuracy. Notably, our method (Llama3.1-8B-instruct + TableDreamer-27K) outperforms R1-Distill-Llama-8B by 3.07, which further validates the effectiveness of the proposed framework and the value of our synthetic data. We believe that TableDreamer framework can be combined with these reasoning LLMs to generate better table instruction tuning data, which could be used to distill more powerful table-related reasoning ability into student LLMs. \textbf{More results and analysis are given in App. \ref{sec:more_results_and_analysis} due to space limitation}, such as the effectiveness on different LLMs and general capacity of different tabular LLMs.

\section{Conclusion}

This paper introduces a novel data synthesis framework for table instruction tuning, which can generate diverse tables together with instructions spanning a wide range of tabular tasks, without relying on any public datasets. At the core of the proposed TableDreamer framework lies the iterative collaboration between input space exploration and  weakness data identification. On the basis of TableDreamer, we construct and release 27K synthetic data, which can effectively enhance LLMs' table understanding ability and outperforms strong baselines. In conclusion, this paper promotes the research of data synthesis for the important table instruction tuning with the new method, dataset and thorough empirical study.

\section{Limitations}
Though this paper presents an effective framework as well as a systematic investigation within the scope of table instruction tuning data synthesis, there are certain limitations and promising directions that deserve future research. First, the proposed framework generates tables and instructions in text format. With the devolvement of multimodal large language models (MLLMs), considerable efforts have been dedicated to the multimodal or visual table understanding problem~\cite{tablellava,table_as_image,TabPedia}, where models take as input a table image rather than a textual table sequence for visual understanding and it also lacks the large amount of diverse instruction tuning data (i.e., triples of table image, instruction, and response). One potential solution is to transform the TableDreamer synthetic textual tables into table images with automatic scripts, e.g., rendering the HTML tables into images with the Python html2image package. 
Second, there are three common paradigms for LLM-based data synthesis: Strong2Weak distillation~\cite{knowledge_distillation_from_strong_teacher}, Weak2Strong Generalization~\cite{weak2strong_generalization}, and self-improving or self-evolving~\cite{self_evolution_llm_survey}. The proposed framework belongs to Strong2Weak distillation paradigm where we use a stronger LLM (Llama3.1-70B-instruct or GPT-4o) to synthesize data in order to enhance the performance of a weaker LLM (e.g., Llama3.1-8B-instruct). The latter two paradigms also require more in-depth explorations, e.g., for the self-evolving paradigm, how can we continuously improve the table understanding ability of the most powerful LLMs like GPT-4o with their own synthetic data.

Third, current data synthesis methods for table understanding and even most table understanding studies, including this paper, are restricted to synthesizing data for the supervised fine-tuning stage. It is worthwhile exploring the generation of table-related preference data to further improve the model performance with reinforce learning~\cite{rDPO,multimodal_peference_data_synthesis}. Particularly, we believe it is a very promising direction to explore incentivizing the table-based Deepseek-R1-style in-depth reasoning~\cite{deepseek_r1} of tabular LLMs by synthesizing tabular task data that can provide reward feedback for reinforcement learning. Lastly, like other data synthesis methods, TableDreamer's synthetic data is not perfect and could contain noisy tables and instruction-response pairs. Valid data filtering strategies would benefit model performance.

\section{Ethical Considerations}
The main objective of the proposed TableDreamer framework is to develop a scalable data synthesis method for table instruction tuning to enhance the table understanding capabilities of LLMs. However, the data generated from the LLMs (Llama3.1-70B-instruct and GPT-4o) may contain harmful content in the synthesized tables, instructions, and responses. To this end, we use the LLM-as-a-judge based on Llama3.1-70B-instruct to check for harmful content within the generated samples, as shown in Fig. \ref{harmful_content_detection_prompt}, and we also randomly sample 5K samples for manually checking. In our empirical evaluations, we do not observe such unsafe data but we still suggest adding relevant safety filtering strategies when using the proposed framework. The benchmarks used in the experiments are free and open datasets for research use, thus the authors foresee no ethical concerns.

\section*{Acknowledgments}
This work was supported by the National Natural Science Foundation of China (No. 62472419, 62472420).

\bibliography{custom}

\newpage
\appendix

\section{More Dataset Information and Comparison}
\label{sec:more_dataset_statistics}
\begin{figure*}[t]
  \centering
  \includegraphics[width=0.85\linewidth]{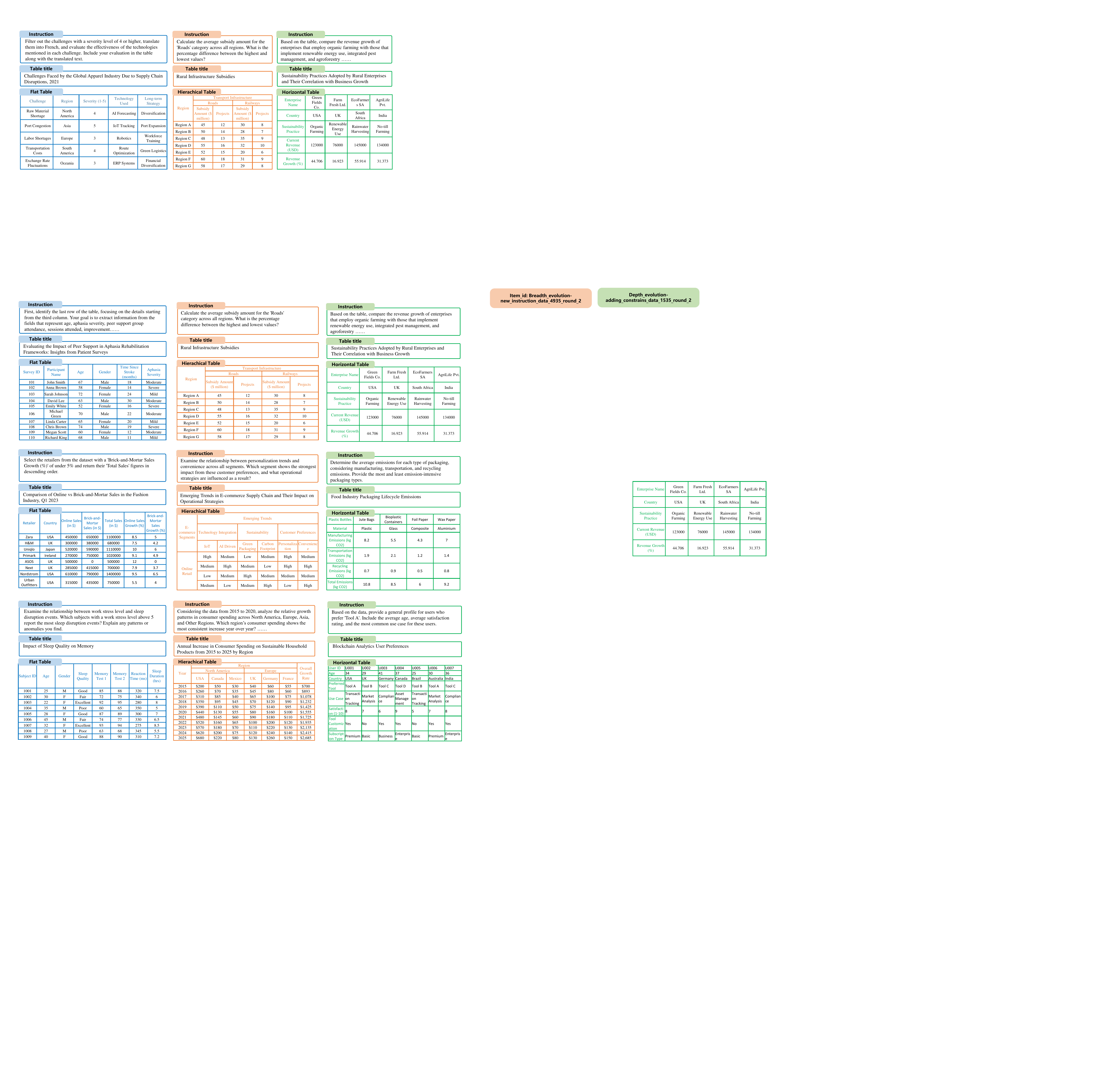}
  \caption{More examples of TableDreamer synthetic data. Tables and instructions are clipped due to space limitation. We render tables into images for better visualization, and real tables could have various formats such as HTML, CSV, Markdown and et al.}
  \label{more_data_samples}
\end{figure*}

Table \ref{tabledreamer_statistics} shows the basic statistics of 27K synthetic data from GPT-4o, such as the average instruction number per table, instruction length (whitespace-split word number) and so on. Figure \ref{more_data_samples} shows more examples of synthetic tables and instructions.

\begin{figure*}[t]
  \centering
  \includegraphics[width=0.8\linewidth]{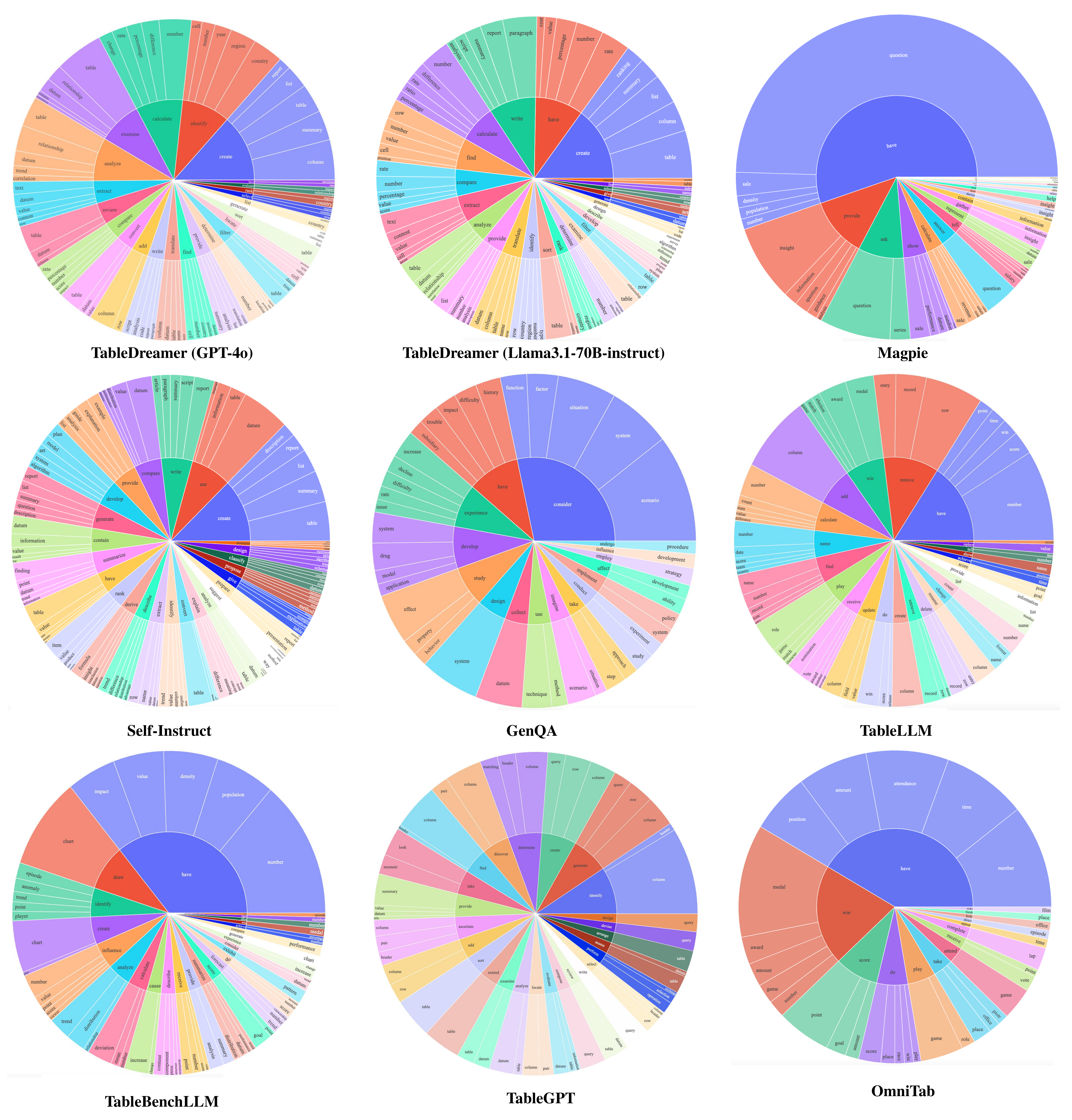}
  \caption{Instruction diversity comparison of different synthetic table instruction tuning data. We show the top 25 most prevalent root verbs (the inner circle) and their top 5 direct nouns (the outer circle) in the synthetic instructions from different methods.}
  \label{inst_verb_comparison}
\end{figure*}

\begin{figure*}[t]
  \centering
  \includegraphics[width=0.8\linewidth]{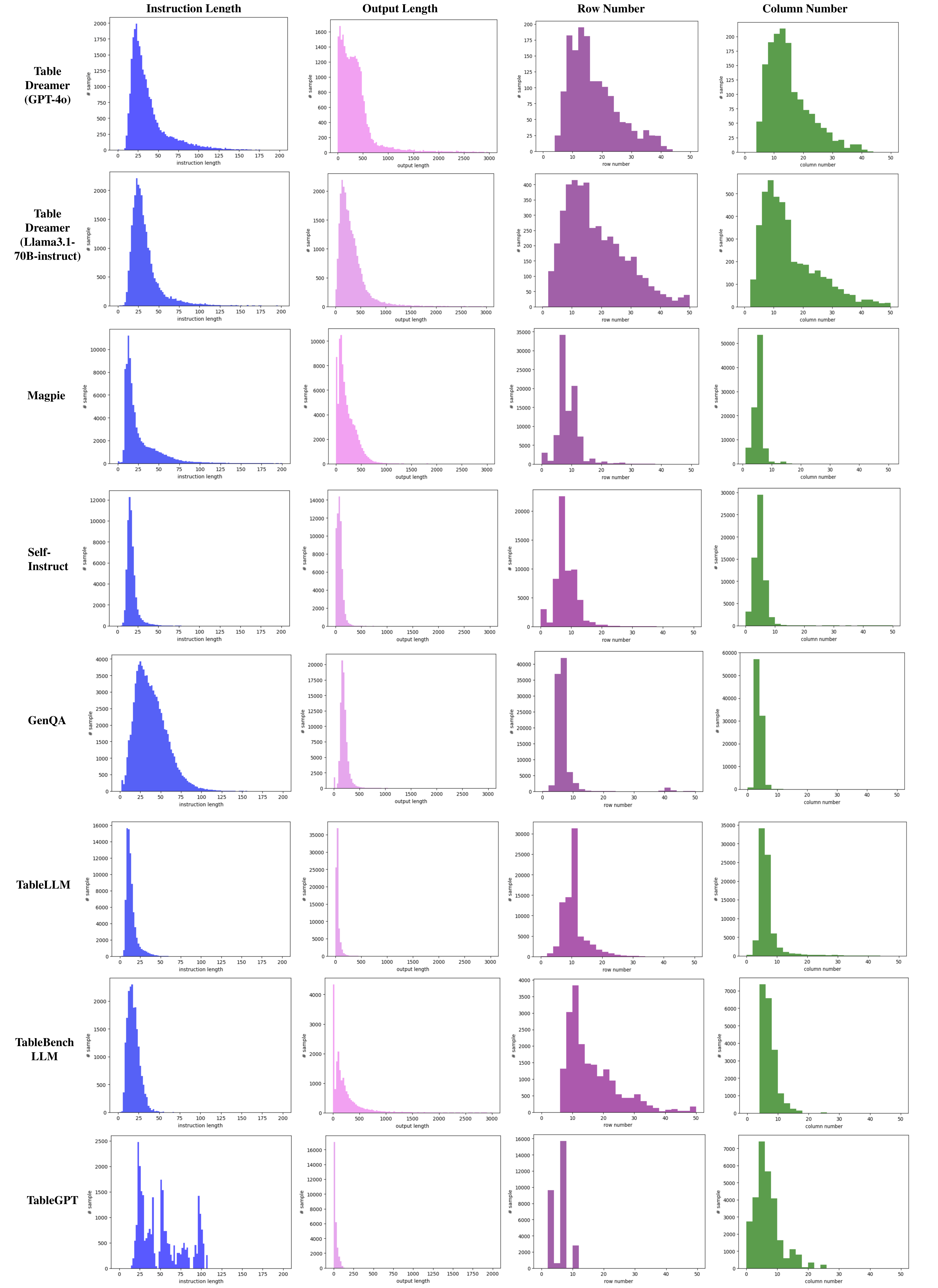}
  \caption{The distribution of instruction length, output length, table row number and table column number in different synthetic table instruction tuning data.}
  \label{data_distribution_comparison}
\end{figure*}

\begin{figure*}[t]
  \centering
  \includegraphics[width=0.7\linewidth]{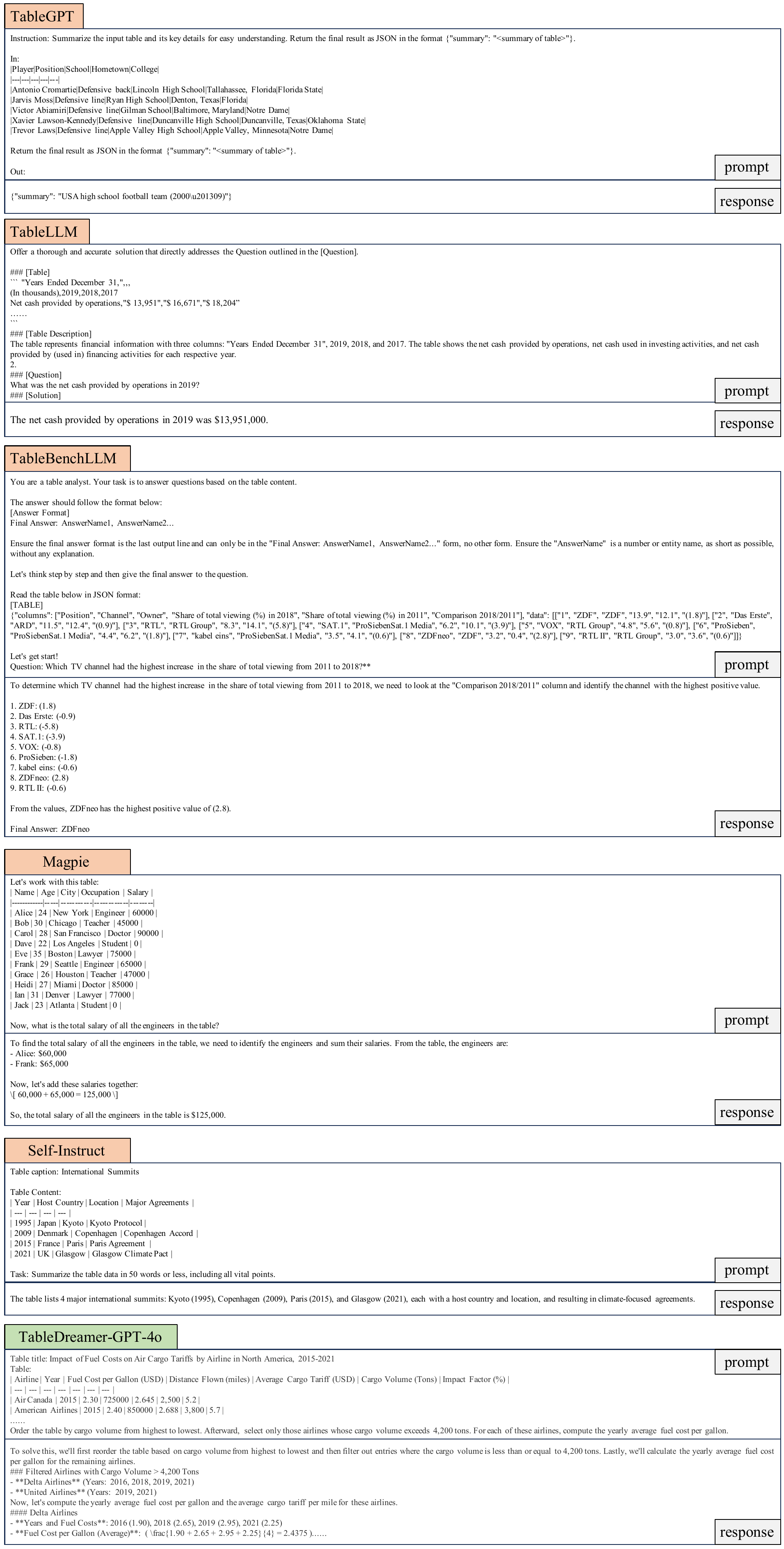}
  \caption{Comparison of synthetic table instruction tuning data from different methods. Some table content are omitted to save space.}
  \label{synthetic_data_example}
\end{figure*}

To shed more light on the characteristics of different data synthesis methods, we visualize the data distribution of various synthetic datasets, including instruction length, output length, table row number, and table column number, and the results are shown in Figure \ref{data_distribution_comparison}. Figure \ref{inst_verb_comparison} further demonstrates the diversity of synthetic instructions from different methods, and Table \ref{table_sft_data_synthesis_method_comparison} compares their characteristics. Figure \ref{synthetic_data_example} directly compares examples from different data synthesis methods. Based on these results, we can find that, although existing data synthesis methods could generate a large amount of table instruction tuning data, the diversity of their synthetic data is limited, e.g., most synthetic samples only cover small tables within 20 rows and 10 columns and can only generate relatively short instructions around 25 words. By contrast, the proposed TableDreamer method offers table instruction tuning data with the best overall diversity.

\section{Implementation Details}
\label{sec:more_implementation_details}

\subsection{TableDreamer}
\label{TableDreamer_implementations}
\begin{figure*}[t]
  \centering
  \includegraphics[width=0.85\linewidth]{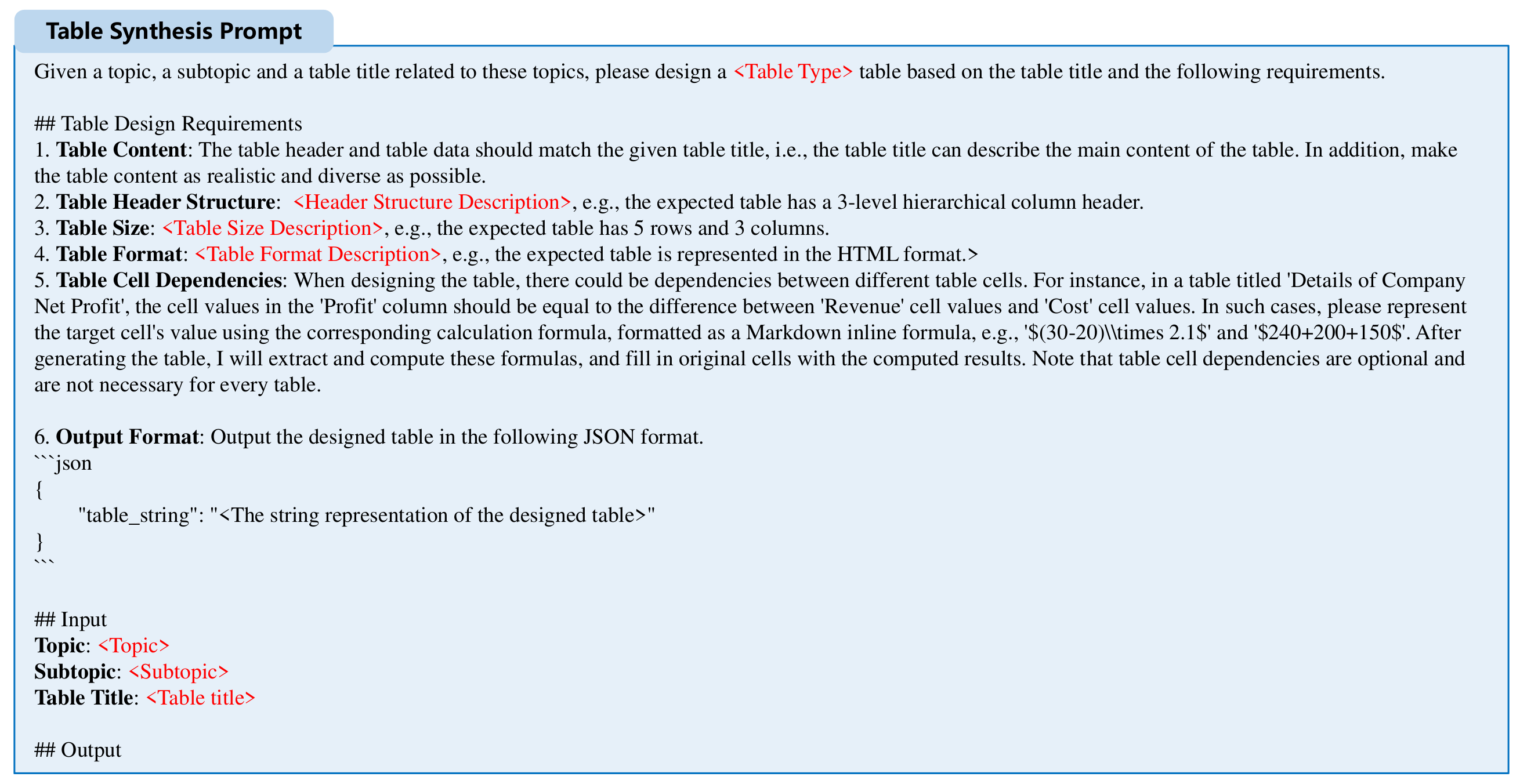}
  \caption{The prompt used for synthesizing diverse tables. The string in red color will be replaced with correlative content in implementation.}
  \label{table_synthesis_prompt}
\end{figure*}

\begin{figure*}[t]
  \centering
  \includegraphics[width=0.85\linewidth]{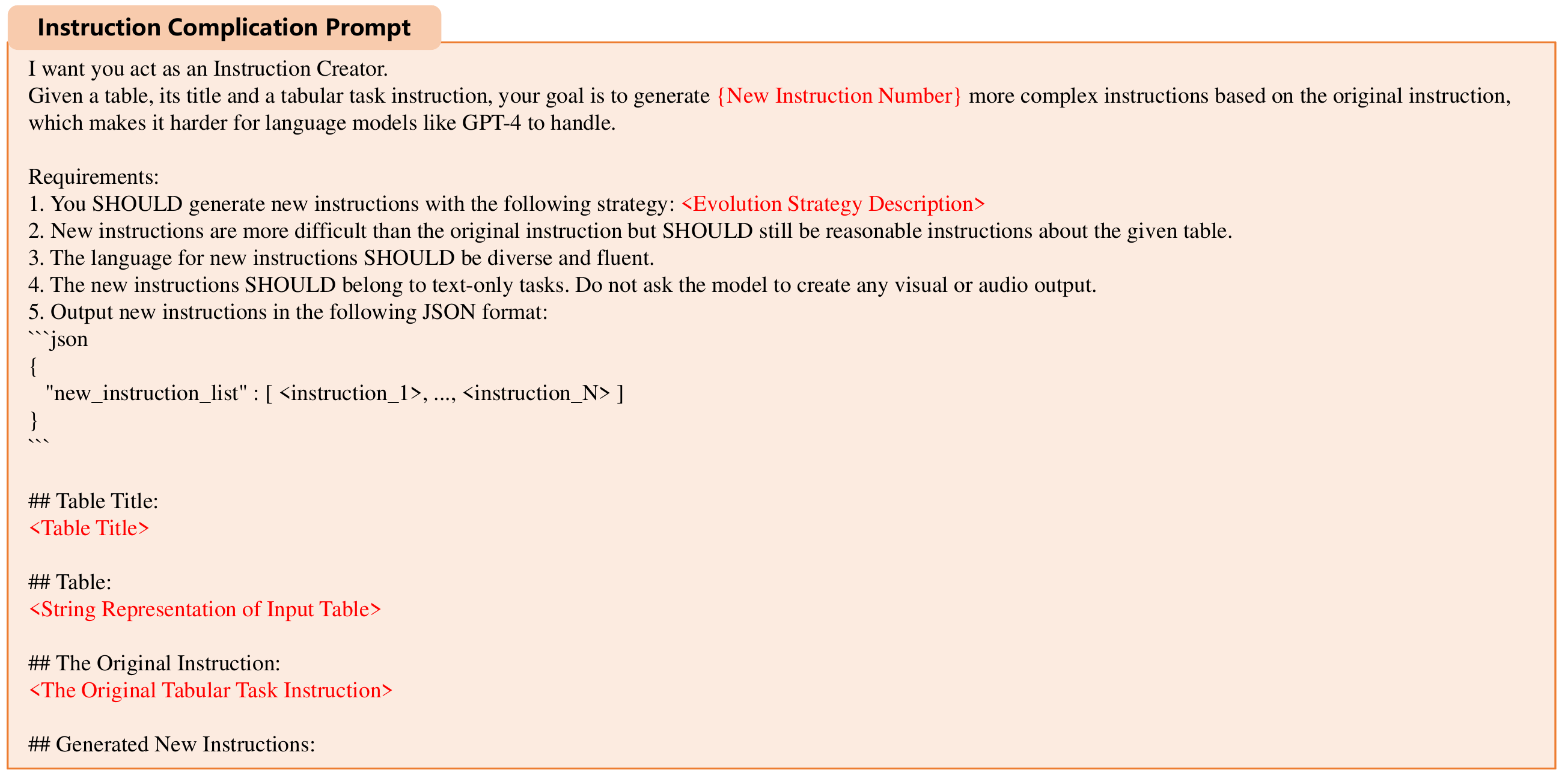}
  \caption{The prompt used for data evolution in the instruction complication direction.}
  \label{inst_complication_prompt}
\end{figure*}

\begin{figure*}[t]
  \centering
  \includegraphics[width=0.85\linewidth]{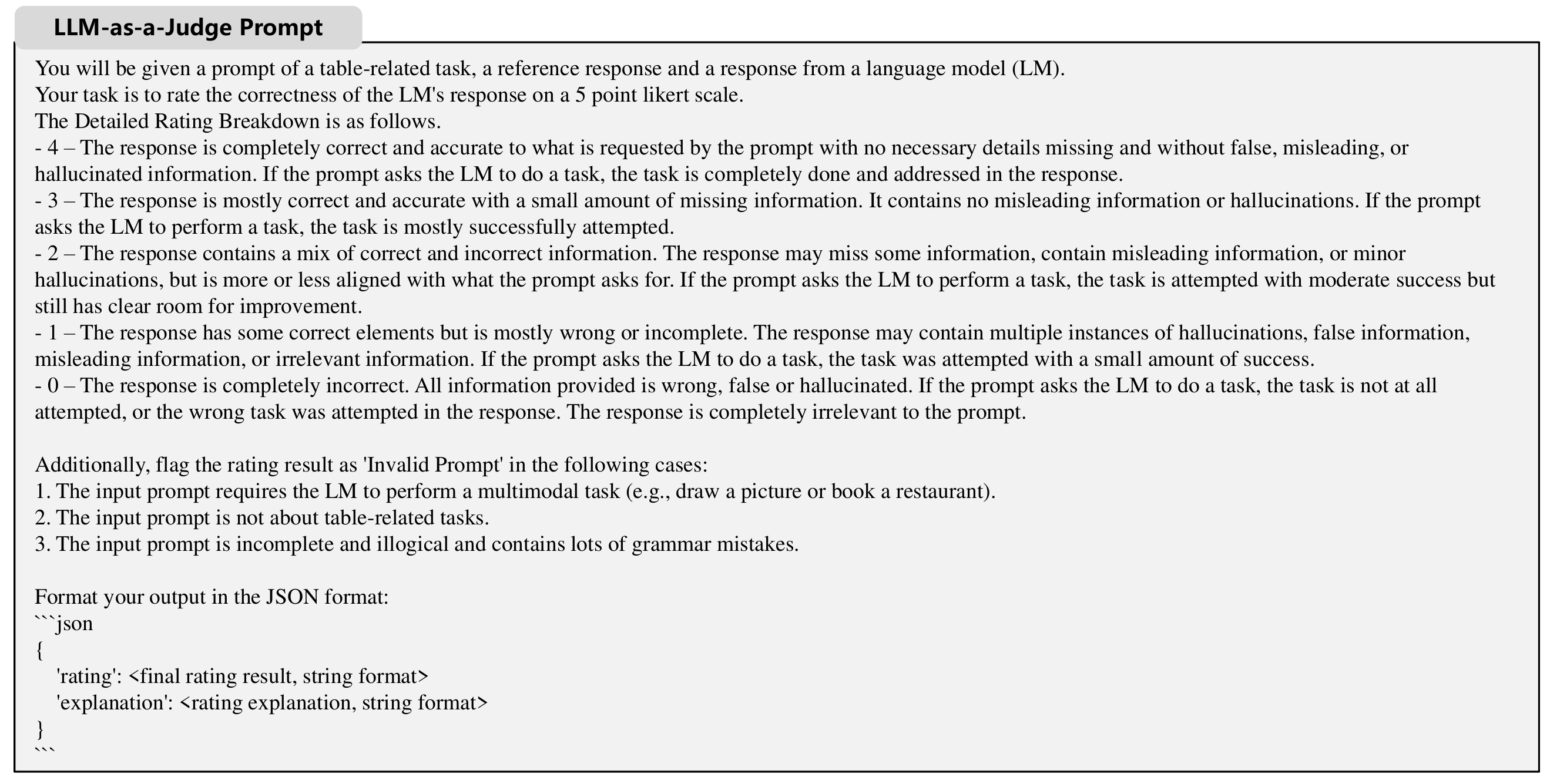}
  \caption{The LLM-as-a-judge prompt used for weakness data identification, which is modified from the correctness judging standard from HelpSteer2~\cite{helpsteer2}.}
  \label{llm_as_a_judge_prompt}
\end{figure*}

\begin{figure*}[t]
  \centering
  \includegraphics[width=0.85\linewidth]{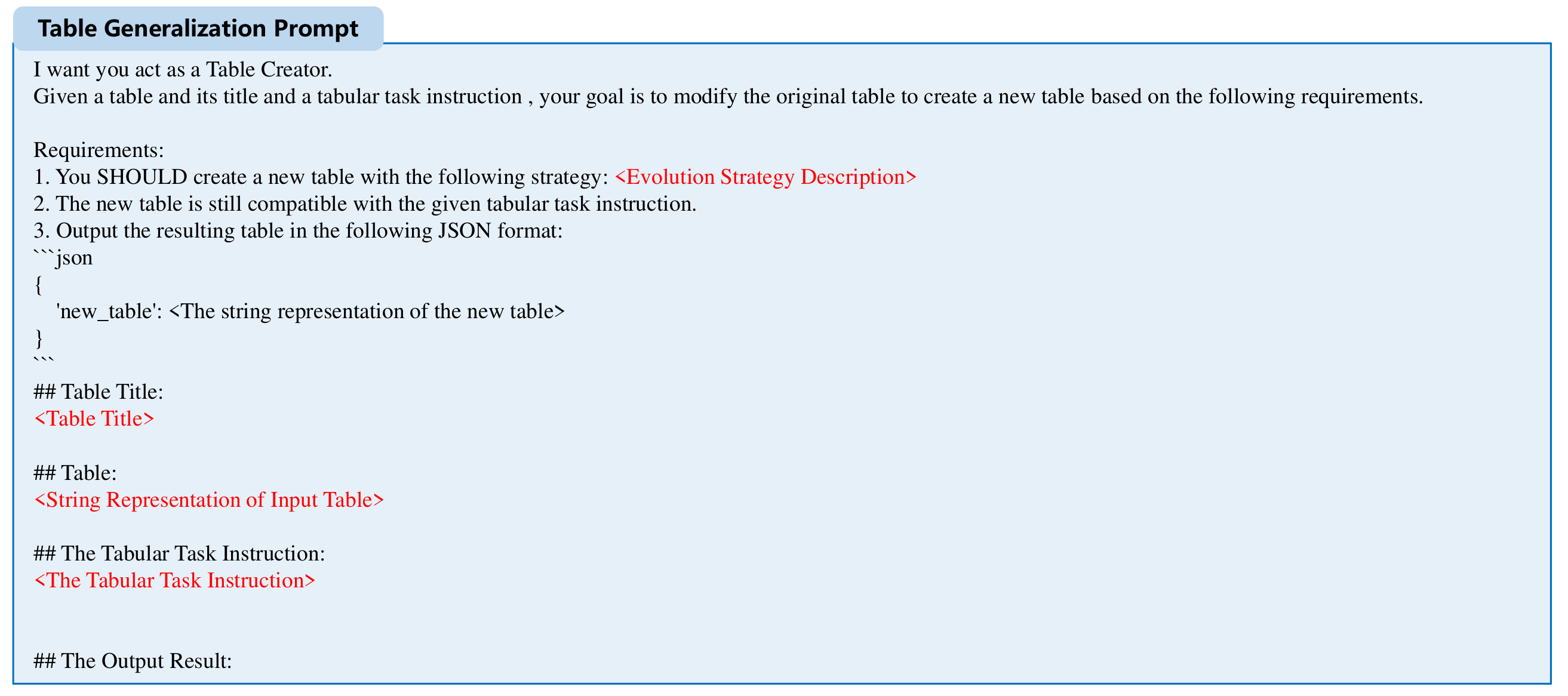}
  \caption{The prompt used for data evolution in the table generalization direction.}
  \label{table_generalization_prompt}
\end{figure*}

\begin{figure*}[t]
  \centering
  \includegraphics[width=0.85\linewidth]{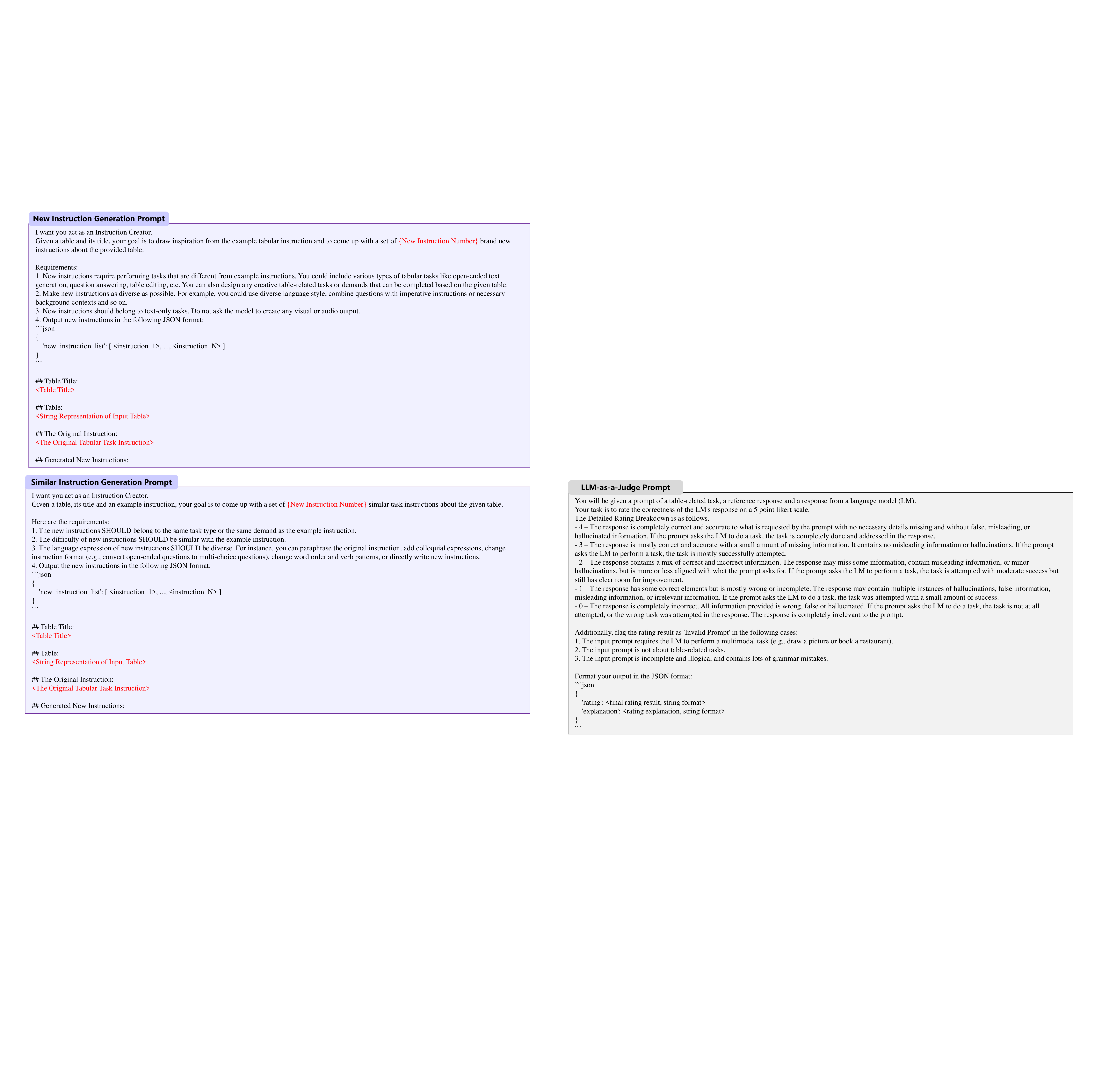}
  \caption{The prompt used for data evolution in the instruction generalization direction.}
  \label{inst_generalization_prompt}
\end{figure*}

The table generation prompt is shown in Figure \ref{table_synthesis_prompt}, and the prompts and strategies for data evolution in three directions are given in Fig. \ref{inst_complication_prompt}, Fig. \ref{inst_generalization_prompt} , Fig. \ref{table_generalization_prompt} and Table \ref{data_evol_strategies}, respectively. The LLM-as-a-judge prompt used for the identification of weaknesses data is shown in Figure \ref{llm_as_a_judge_prompt}, which is modified from the correctness-judging standard of the HelpSteer2 dataset~\cite{helpsteer2}. The 20 seed tasks and their descriptions are given in Table \ref{seed_task_descriptions}, which are used by the teacher LLM (e.g., Llama3.1-70B-instruct or GPT-4o) to generate seed instructions based on synthetic tables. Multiple prompt templates are used to combine the input table, table title and instruction to form the final input prompt in the training data. During fine-tuning, we adopt the recommended hyper-parameters from \citet{rethinking_table_instruction_tuning} and perform the standard supervised fine-tuning with a learning rate of 1e-6 and a batch size of 128 for 2 epochs. During inference, we set the temperature to 0.01 for reproducible evaluation results.

\begin{table}[t]\footnotesize
\centering
\renewcommand{\arraystretch}{1.3}
\setlength\tabcolsep{2pt}
\scalebox{0.8}{
\begin{tabular}{lc} 
\hline
\textbf{Characteristic} & \textbf{Value} \\ 
\hline
Avg. instruction number per table & 3.4 \\
Row number per table (median/mean/min/max) & 15/16.8/4/43 \\
Column number per table (median/mean/min/max) & 13/14.8/4/45 \\
Cell number per table (median/mean/min/max) & 200/237/28/1008 \\
Instruction length by word (median/mean/min/max) & 29/36.9/6/900 \\
Output length by word (median/mean/min/max) & 288/412.9/3/11513 \\
\hline
\end{tabular}

}
\caption{Basic statistics of the TableDreamer-27K synthetic data generated by GPT-4o.} 
\label{tabledreamer_statistics}
\end{table}

\begin{table*}[t]\footnotesize
\centering
\renewcommand{\arraystretch}{1.3}
\setlength\tabcolsep{2pt}
\scalebox{0.83}{
\begin{tabular}{c|c|ccccc|cc|cc|c} 
\hline
\multirow{2}{*}{\textbf{Methods}} & \multirow{2}{*}{\textbf{LLMs}} & \multicolumn{5}{c|}{\textbf{TQA }} & \multicolumn{2}{c|}{\textbf{TFV }} & \multicolumn{2}{c|}{\textbf{T2T }} & \multirow{2}{*}{\textbf{Ave. Acc}} \\ 
\cline{3-11}
 &  & \textbf{TABMWP} & \textbf{WTQ} & \textbf{HiTab} & \textbf{AIT-QA} & \textbf{TabMCQ} & \textbf{TabFact} & \textbf{InfoTabs} & \textbf{FeTaQA} & \textbf{QTSumm} &  \\ 
\hline
\multirow{2}{*}{Reasoning LLMs} & DeepSeek-R1 (671B) & 64.75 & \textbf{77.76} & \textbf{39.34} & 65.36 & \textbf{88.38} & 73.64 & 69.59 & \textbf{97.20} & \textbf{94.06} & 74.45 \\
 & QwQ-32B & \textbf{65.46} & 77.07 & 36.14 & \textbf{76.12} & 87.85 & \textbf{78.88} & \textbf{74.74} & 95.16 & 92.39 & \textbf{75.98} \\ 
\hline
\multirow{2}{*}{\begin{tabular}[c]{@{}c@{}}Distilled Reasoning \\LLMs\end{tabular}} & R1-Distill-Llama-70B & 63.10 & 62.26 & 19.61 & 56.36 & 86.68 & 68.89 & 72.53 & 93.51 & 90.82 & 68.20 \\
 & R1-Distill-Llama-8B & 63.50 & 46.82 & 18.02 & 47.55 & 82.02 & 52.57 & 60.38 & 86.02 & 76.16 & 59.23 \\ 
\hline
\multirow{3}{*}{Traditional LLMs} & DeepSeek-V3 (671B) & 70.29 & 65.65 & 32.86 & 63.79 & 89.31 & 65.46 & 67.14 & 96.21 & 90.54 & 71.25 \\
 & GPT-4o & 73.47 & 68.57 & 38.13 & 71.62 & 88.04 & 69.03 & 68.11 & 95.66 & 92.30 & 73.88 \\
 & Llama3.1-70B-instruct & 61.69 & 50.43 & 19.35 & 61.25 & 86.88 & 66.31 & 62.55 & 90.61 & 84.04 & 64.79 \\ 
\hline
\multirow{3}{*}{Ours} & Llama3.1-8B-instruct & 53.39 & 36.53 & 11.35 & 43.63 & 75.31 & 53.87 & 48.94 & 78.98 & 66.98 & 52.11 \\
 & w/ TableDreamer-27K & 64.61 & 54.66 & 22.88 & 53.22 & 84.29 & 63.09 & 57.65 & 84.37 & 75.97 & 62.30 \\
 & $\triangle$ & 1.11 & 7.84 & 4.86 & 5.67 & 2.27 & 10.52 & -2.73 & -1.65 & -0.19 & 3.07 \\
\hline
\end{tabular}

}
\caption{Results of recent R1-style reasoning LLMs. $\triangle$ indicates performance increase between our method and R1-Distill-Llama-8B.} 
\label{reasoning_LLM_results}
\end{table*}

\begin{table}[t]\footnotesize
\centering
\renewcommand{\arraystretch}{1.3}
\setlength\tabcolsep{2pt}
\scalebox{0.82}{
\begin{tabular}{c|cccc|c} 
\hline
\textbf{Method} & \textbf{TQA} & \textbf{TFV} & \textbf{T2T} & \textbf{TableGPT} & \textbf{Ave. Acc} \\ 
\hline
Llama3.1-8B-Instruct & 44.04 & 51.41 & 72.98 & 21.68 & 49.07 \\
w/ TableDreamer-27K & 55.93 & 60.37 & 80.17 & 46.20 & 60.69 \\
$\triangle$ & {\cellcolor[rgb]{0.741,1,0.741}}11.89 & {\cellcolor[rgb]{0.792,1,0.792}}8.96 & {\cellcolor[rgb]{0.792,1,0.792}}7.19 & {\cellcolor[rgb]{0.741,1,0.741}}24.52 & {\cellcolor[rgb]{0.741,1,0.741}}11.63 \\ 
\hline
Mistral-7B-Instruct-v0.3 & 37.98 & 45.24 & 67.10 & 44.26 & 45.88 \\
w/ TableDreamer-27K & 51.06 & 49.34 & 76.19 & 43.33 & 54.97 \\
$\triangle$ & {\cellcolor[rgb]{0.741,1,0.741}}13.07 & {\cellcolor[rgb]{0.906,1,0.906}}4.10 & {\cellcolor[rgb]{0.792,1,0.792}}9.09 & {\cellcolor[rgb]{1,0.792,0.792}}-0.93 & {\cellcolor[rgb]{0.792,1,0.792}}9.08 \\ 
\hline
MiniCPM3-4B & 46.77 & 35.21 & 55.47 & 40.79 & 45.60 \\
w/ TableDreamer-27K & 53.03 & 40.50 & 64.51 & 42.85 & 51.80 \\
$\triangle$ & {\cellcolor[rgb]{0.792,1,0.792}}6.27 & {\cellcolor[rgb]{0.792,1,0.792}}5.29 & {\cellcolor[rgb]{0.792,1,0.792}}9.04 & {\cellcolor[rgb]{0.906,1,0.906}}2.06 & {\cellcolor[rgb]{0.792,1,0.792}}6.20 \\ 
\hline
InternLM2.5-7B-Chat & 36.81 & 46.20 & 72.15 & 39.52 & 46.02 \\
w/ TableDreamer-27K & 54.99 & 51.99 & 73.08 & 40.14 & 56.52 \\
$\triangle$ & {\cellcolor[rgb]{0.741,1,0.741}}18.18 & {\cellcolor[rgb]{0.792,1,0.792}}5.79 & {\cellcolor[rgb]{0.906,1,0.906}}0.93 & {\cellcolor[rgb]{0.906,1,0.906}}0.62 & {\cellcolor[rgb]{0.741,1,0.741}}10.50 \\
\hline
\end{tabular}

}
\caption{Comparison of average performance of different LLMs fine-tuned with the TableDreamer data.} 
\label{different_LLM_results}
\end{table}

\begin{table}[t]\footnotesize
\centering
\renewcommand{\arraystretch}{1.3}
\setlength\tabcolsep{2pt}
\scalebox{0.87}{
\begin{tabular}{c|cccc|c} 
\hline
\textbf{\textbf{Method}} & \textbf{TQA} & \textbf{TFV} & \textbf{T2T} & \textbf{TableGPT} & \textbf{Ave. Acc} \\ 
\hline
Llama3.1-8B-Instruct & 44.04 & 51.41 & 72.98 & 21.68 & 49.07 \\
w/ TableDreamer-27K & 55.93 & 60.37 & 80.17 & 46.20 & 60.69 \\
w/o Weakness Iden.-34K & 53.12 & 51.72 & 75.82 & 42.12 & 56.28 \\
$\triangle$ & {\cellcolor[rgb]{1,0.741,0.741}}2.81 & {\cellcolor[rgb]{1,0.675,0.675}}8.65 & {\cellcolor[rgb]{1,0.741,0.741}}4.36 & {\cellcolor[rgb]{1,0.741,0.741}}4.08 & {\cellcolor[rgb]{1,0.741,0.741}}4.41 \\ 
\hline
Mistral-Instruct-v0.3-7B & 37.98 & 45.24 & 67.10 & 44.26 & 45.88 \\
w/ TableDreamer-27K & 51.06 & 49.34 & 76.19 & 43.33 & 54.97 \\
w/o Weakness Iden.-34K & 49.87 & 48.17 & 72.81 & 45.29 & 53.66 \\
$\triangle$ & {\cellcolor[rgb]{1,0.835,0.835}}1.19 & {\cellcolor[rgb]{1,0.835,0.835}}1.17 & {\cellcolor[rgb]{1,0.741,0.741}}3.38 & {\cellcolor[rgb]{0.741,1,0.741}}+1.96 & {\cellcolor[rgb]{1,0.835,0.835}}1.31 \\ 
\hline
MiniCPM3-4B & 46.77 & 35.21 & 55.47 & 40.79 & 45.60 \\
w/ TableDreamer-27K & 53.03 & 40.50 & 64.51 & 42.85 & 51.80 \\
w/o Weakness Iden.-34K & 52.05 & 39.71 & 64.40 & 41.81 & 51.03 \\
$\triangle$ & {\cellcolor[rgb]{1,0.835,0.835}}0.99 & {\cellcolor[rgb]{1,0.835,0.835}}0.79 & {\cellcolor[rgb]{1,0.835,0.835}}0.11 & {\cellcolor[rgb]{1,0.835,0.835}}1.04 & {\cellcolor[rgb]{1,0.835,0.835}}0.78 \\ 
\hline
InternLM2.5-7B-Chat & 36.81 & 46.20 & 72.15 & 39.52 & 46.02 \\
w/ TableDreamer-27K & 54.99 & 51.99 & 73.08 & 40.14 & 56.52 \\
w/o Weakness Iden.-34K & 50.78 & 49.01 & 71.24 & 45.14 & 53.96 \\
$\triangle$ & {\cellcolor[rgb]{1,0.741,0.741}}4.20 & {\cellcolor[rgb]{1,0.741,0.741}}2.98 & {\cellcolor[rgb]{1,0.835,0.835}}1.84 & {\cellcolor[rgb]{0.741,1,0.741}}+5.00 & {\cellcolor[rgb]{1,0.741,0.741}}2.57 \\
\hline
\end{tabular}
}
\caption{Ablation study on different LLMs fine-tuned with the TableDreamer-34K data without weakness identification.} 
\label{different_LLM_results_without_weakness_idtentification}
\end{table}

\begin{table}[t]\footnotesize
\centering
\renewcommand{\arraystretch}{1.3}
\setlength\tabcolsep{2pt}
\scalebox{0.87}{
\begin{tabular}{c|cccc|c} 
\hline
\textbf{Method} & \textbf{TQA} & \textbf{TFV} & \textbf{T2T} & \textbf{TableGPT} & \textbf{Ave. Acc} \\ 
\hline
Llama3.1-8B-instruct & 44.04 & 51.41 & 72.98 & 21.68 & 49.07 \\
w/ TableDreamer-27K & 52.01 & 53.65 & 79.90 & 33.03 & 56.02 \\
\begin{tabular}[c]{@{}c@{}}w/ randomly selected \\weakness data-27K\end{tabular} & 48.54 & 50.58 & 76.72 & 30.06 & 52.73 \\
$\triangle$ & {\cellcolor[rgb]{1,0.741,0.741}}3.47 & {\cellcolor[rgb]{1,0.741,0.741}}3.07 & {\cellcolor[rgb]{1,0.741,0.741}}3.18 & {\cellcolor[rgb]{1,0.741,0.741}}2.97 & {\cellcolor[rgb]{1,0.741,0.741}}3.29 \\
\hline
\end{tabular}
}
\caption{Ablation study of randomly selecting weakness data for data evolution with Llama3.1-70B-Instruct.} 
\label{ablation_study_of_randomly_selected_weakness_data}
\end{table}

\begin{table*}[t]\footnotesize
\centering
\renewcommand{\arraystretch}{1.3}
\setlength\tabcolsep{3pt}
\scalebox{0.75}{
\begin{tabular}{c|c|c|c|l|c} 
\hline
\textbf{Task Category} & \textbf{Benchmark} & \textbf{\# Test samples} & \textbf{Ave. Input Length} & \multicolumn{1}{c|}{\textbf{Task Description}} & \textbf{Metric} \\ 
\hline
\multirow{5}{*}{\begin{tabular}[c]{@{}c@{}}Table \\Question \\Answering \\(TQA)\end{tabular}} & WTQ & 4344 & 496.3 & \begin{tabular}[c]{@{}l@{}}TQA based on tables which usually possesses a flat\\structure with the first row as the sole column header.\end{tabular} & Accuracy \\ 
\cline{2-6}
 & HiTab & 1576 & 399.4 & \begin{tabular}[c]{@{}l@{}}TQA based on tables which usually possesses\\ hierachical headers and merged cells.\end{tabular} & Accuracy \\ 
\cline{2-6}
 & AIT-QA & 511 & 275.2 & TQA based on hierarchical tables from the airline industry. & Accuracy \\ 
\cline{2-6}
 & TabMCQ & 1029 & 311.8 & TQA with multi-choice questions. & Accuracy \\ 
\cline{2-6}
 & TABMWP & 7686 & 89.6 & \begin{tabular}[c]{@{}l@{}}TQA requiring mathematical reasoning operations such as\\finding the largest number or do math computations.\end{tabular} & Accuracy \\ 
\hline
\multirow{2}{*}{\begin{tabular}[c]{@{}c@{}}Table \\Fact\\Verification \\(TFV)\end{tabular}} & TabFact & 6845 & 303.7 & \begin{tabular}[c]{@{}l@{}}Given a table as evidence and a statement, the\\ task is to distinguish whether the given\\ statement is entailed or refuted by the table.\end{tabular} & Accuracy \\ 
\cline{2-6}
 & InfoTabs & 5400 & 155.1 & \begin{tabular}[c]{@{}l@{}}Given a infobox table as evidence and a statement, \\the task is to distinguish whether the \\givenstatement is entailed or refuted by the table.\end{tabular} & Accuracy \\ 
\hline
\multirow{2}{*}{\begin{tabular}[c]{@{}c@{}}Table to \\Text \\Generation (T2T)\end{tabular}} & QTSumm & 1078 & 242.8 & \begin{tabular}[c]{@{}l@{}}Given a table and a query, models must perform \\human-like reasoning and analysis over \\the given table to generate a tailored summary.\end{tabular} & \begin{tabular}[c]{@{}c@{}}LLM-as-a-judge \\Acc.\end{tabular} \\ 
\cline{2-6}
 & FeTaQA & 2003 & 263 & \begin{tabular}[c]{@{}l@{}}TQA with a free-form text answer rather \\than a~short text span copied from the table.\end{tabular} & \begin{tabular}[c]{@{}c@{}}LLM-as-a-judge \\Acc.\end{tabular} \\ 
\hline
\multirow{5}{*}{TableGPT} & \begin{tabular}[c]{@{}c@{}}Column \\Finding\end{tabular} & 1682 & 106.3 & \begin{tabular}[c]{@{}l@{}}Identify the column-name of a \\specific value that appears only once in a given table\end{tabular} & Accuracy \\ 
\cline{2-6}
 & \begin{tabular}[c]{@{}c@{}}Data \\Imputation\end{tabular} & 2000 & 147.8 & \begin{tabular}[c]{@{}l@{}}Predict the missing values in a cell \\based on the table context\end{tabular} & Accuracy \\ 
\cline{2-6}
 & \begin{tabular}[c]{@{}c@{}}Row2Row \\Transformation\end{tabular} & 570 & 101.7 & \begin{tabular}[c]{@{}l@{}}Transform table data based \\on input/output examples\end{tabular} & Accuracy \\ 
\cline{2-6}
 & \begin{tabular}[c]{@{}c@{}}Missing Value \\Identification\end{tabular} & 8000 & 107.1 & \begin{tabular}[c]{@{}l@{}}Identify the row and column \\position of the only missing cell in a given table\end{tabular} & Accuracy \\ 
\cline{2-6}
 & \begin{tabular}[c]{@{}c@{}}TQA\\(SQA,WTQ)\end{tabular} & 9048 & 229.5 & \begin{tabular}[c]{@{}l@{}}Answer a natural-language question \\based on the content of a table\end{tabular} & Accuracy \\
\hline
\end{tabular}
}

\caption{Detailed description and statistics of 10 used benchmarks. The average input length is computed by whitespace-split word number.}
\label{benchmark_descriptions}
\end{table*}

\begin{table*}
\centering
\resizebox{\linewidth}{!}{
\begin{tabular}{c|cccccccc} 
\hline
\textbf{Method} & \begin{tabular}[c]{@{}c@{}}\textbf{Rely on }\\\textbf{public tables?}\end{tabular} & \begin{tabular}[c]{@{}c@{}}\textbf{Need human }\\\textbf{annotators?}\end{tabular} & \textbf{Table types} & \textbf{Table formats} & \begin{tabular}[c]{@{}c@{}}\textbf{Instruction }\\\textbf{template}\end{tabular} & \begin{tabular}[c]{@{}c@{}}\textbf{Instruction generation }\\\textbf{method}\end{tabular} & \begin{tabular}[c]{@{}c@{}}\textbf{Response }\\\textbf{type}\end{tabular} & \begin{tabular}[c]{@{}c@{}}\textbf{Consider model }\\\textbf{weakness?}\end{tabular} \\ 
\hline
OmniTab & Yes & No & Flat & Python-dict-style & Fixed & SQL2NL & Short answer & No \\
ReasTap & Yes & No & Flat & Python-dict-style & Fixed & Predefined template & Short answer & No \\
UCTR & Yes & No & Flat & Python-dict-style & Fixed & Program2NL & Short answer & No \\
TableLlama & Yes & No & \begin{tabular}[c]{@{}c@{}}Flat, \\hierarchical\end{tabular} & heuristically-defined & Fixed & Converting existing datasets & Short answer & No \\
TableGPT & Yes & No & Flat & Markdown-style & Fixed & LLM generated & Short answer & No \\
TableBenchLLM & Yes & Yes & \begin{tabular}[c]{@{}c@{}}Flat, \\hierarchical\end{tabular} & Python-dict-style & Fixed & LLM generated+Human Corrected & Detail Reasoning steps & No \\
TableLLM & Yes & No & Flat & CSV & Fixed & LLM generated & Detail Reasoning steps & No \\ 
\hline
\textbf{TableDreamer (Ours)} & \textbf{No} & \textbf{No} & \begin{tabular}[c]{@{}c@{}}\textbf{Flat, hierarchical, }\\\textbf{horizontal}\end{tabular} & \textbf{Diversified} & \textbf{Diversified} & \textbf{LLM generated} & \textbf{Detail Reasoning steps} & \textbf{Yes} \\
\hline
\end{tabular}
}
\caption{Comparison of TableDreamer and previous table instruction tuning data synthesis methods.}
\label{table_sft_data_synthesis_method_comparison}
\end{table*}

\subsection{Baseline Implementations.}
\label{baseline_implementations}

For general LLMs and tabular LLMs, we directly evaluate their performance on the collected benchmarks using model checkpoints downloaded from the HuggingFace. For general data synthesis baselines like Self-Instruct, we made necessary adjustments to enable them to generate table instruction tuning data based on the Llama3.1-70B-instruct. For table instruction tuning data generation baselines, we directly use their released synthetic data as the training data. We fine-tune the Llama3.1-8B-instruct with the synthetic data from the TableDreamer and other data synthesis baselines, and evaluate the performance of the resulting models to compare the effectiveness of different data synthesis methods.

Here we give more details about reimplementing general data synthesis baselines. For Self-Instruct~\cite{wang-etal-2023-self-instruct}, we construct 175 general tabular task requests with the help of GPT-4o and use them as seed data to generate more tabular tasks with the original self-instruct framework. Then, the filtered tabular tasks are used to synthesize task-inputs which include input tables and instructions. For Magpie~\cite{magpie} reimplementation, we follow the original Magpie approach which modifies the system prompt to generate domain-specific instruction tuning like mathematical data. Similarly, we modify the system prompt to ask the LLM to act as a table understanding expert that fulfills table-related tasks. Then, a pre-query template with the modified system prompt is input to the LLM and it will autonomously generate the input table and related instructions autoregressively, which are further filtered with the methods from the original paper. 

The GenQA~\cite{GenQA} explores different prompts to synthesize instruction-tuning data. To reimplement GenQA, we modify the Generator-Conditional data synthesis prompt, the best prompt according to the paper, to generate input tables and instructions based on the diverse topics from the original paper. For Evol-Instruct~\cite{xu2023wizardlm}, we select 40K samples generated from Magpie and Self-Instruct synthetic data as seed data for synthesizing new samples with the evol-instruct prompts. The Dynosaur~\cite{yin-etal-2023-dynosaur} method synthesizes instruction-tuning data by transforming existing datasets with LLM-generated instructions. To reimplement Dynosaur, we collect 5 table understanding datasets including FinQA, SQA, WikiSQL, TAT-QA and PubHealthTab as the basic data source and carefully construct their dataset metadata, which are used by the Llama3.1-70B-instruct to design tabular tasks and instructions. More details and introduction about these baselines could be found in the original papers. All experiments in this paper were conducted on one machine with 8 80GB A100.

\begin{table*}[t]\footnotesize
\centering
\renewcommand{\arraystretch}{1.3}
\setlength\tabcolsep{3pt}
\scalebox{0.70}{
\begin{tabular}{c|c|l} 
\hline
Task Category & Task Name & \multicolumn{1}{c}{Task Description} \\ 
\hline
\multirow{5}{*}{\begin{tabular}[c]{@{}c@{}}Table \\Question \\Answering\end{tabular}} & \begin{tabular}[c]{@{}c@{}}Numerical\\reasoning problem\end{tabular} & \begin{tabular}[c]{@{}l@{}}Given a table and a problem, the model needs to perform mathematical calculations based on numerical values in \\the table and the problem, such as addition, subtraction, averaging, calculation of growth rates, etc.\end{tabular} \\ 
\cline{2-3}
 & \begin{tabular}[c]{@{}c@{}}Information\\seeking problem\end{tabular} & \begin{tabular}[c]{@{}l@{}}Given a table and a related problem, the model needs to conduct information seeking from \\table cells based on the requirements of problem.\end{tabular} \\ 
\cline{2-3}
 & \begin{tabular}[c]{@{}c@{}}Multihop \\reasoning problem\end{tabular} & \begin{tabular}[c]{@{}l@{}}Given a table and a related problem, the model needs to conduct multi-hop reasoning according\\~to the requirements of the problem to get the final answer.\end{tabular} \\ 
\cline{2-3}
 & \begin{tabular}[c]{@{}c@{}}Time \\calculation problem\end{tabular} & \begin{tabular}[c]{@{}l@{}}Given a table and a problem, the model needs to perform temporal calculations or comparison based on\\~the time information, such as calculating the time difference between the release time of two movies.\end{tabular} \\ 
\cline{2-3}
 & \begin{tabular}[c]{@{}c@{}}Table-based \\fact verification\end{tabular} & \begin{tabular}[c]{@{}l@{}}Given a table and a statement, the model needs to determine whether the statement is true based on \\the table information.~\end{tabular} \\ 
\hline
\multirow{3}{*}{\begin{tabular}[c]{@{}c@{}}Table-to-text \\generation\end{tabular}} & \begin{tabular}[c]{@{}c@{}}Table \\description\end{tabular} & Given a table, the model needs to describe the table contents in detail. \\ 
\cline{2-3}
 & \begin{tabular}[c]{@{}c@{}}Table \\summarization\end{tabular} & \begin{tabular}[c]{@{}l@{}}Given a table, the model is asked to summarize the key information in the table \\and generate a summary.\end{tabular} \\ 
\cline{2-3}
 & \begin{tabular}[c]{@{}c@{}}Table \\analysis\end{tabular} & \begin{tabular}[c]{@{}l@{}}Given a table, the model is asked to act as a professional data analyst, analyzing the key trends and \\phenomena in the table data, such as analyzing the sales of products in each quarter against the sales report.\end{tabular} \\ 
\hline
\multirow{5}{*}{\begin{tabular}[c]{@{}c@{}}Table \\Structure \\Understanding\end{tabular}} & \begin{tabular}[c]{@{}c@{}}Table size \\detection\end{tabular} & Given a table, the model is asked to determine how many rows and columns the table has. \\ 
\cline{2-3}
 & \begin{tabular}[c]{@{}c@{}}Table cell \\extraction\end{tabular} & \begin{tabular}[c]{@{}l@{}}Given a table and some cell locations (represented by row and column numbers), \\the model is asked to extract the cell text for the corresponding location.\end{tabular} \\ 
\cline{2-3}
 & \begin{tabular}[c]{@{}c@{}}Table \\cell location\end{tabular} & \begin{tabular}[c]{@{}l@{}}Given a table and some cell text, the model is asked to find the position of those cells in the table\\~(represented by row and column numbers).\end{tabular} \\ 
\cline{2-3}
 & \begin{tabular}[c]{@{}c@{}}Row\&Column \\extraction\end{tabular} & \begin{tabular}[c]{@{}l@{}}Given a table and some row or column numbers, the model is asked to extract all the text for the \\corresponding row or column.\end{tabular} \\ 
\cline{2-3}
 & \begin{tabular}[c]{@{}c@{}}Merged \\cell detection\end{tabular} & \begin{tabular}[c]{@{}l@{}}Given a table, the model is asked to determine whether~the table contains merged cells and give the location of \\all the merged cells (represented by row and column numbers) if so.\end{tabular} \\ 
\hline
\multirow{5}{*}{\begin{tabular}[c]{@{}c@{}}Data \\Manipulation\end{tabular}} & \begin{tabular}[c]{@{}c@{}}Data \\formating\end{tabular} & \begin{tabular}[c]{@{}l@{}}Given a table and user requirements, the model needs to modify the formats of some table data according \\to user requirements.\end{tabular} \\ 
\cline{2-3}
 & \begin{tabular}[c]{@{}c@{}}Data \\cleaning\end{tabular} & \begin{tabular}[c]{@{}l@{}}Given a table that may contain noise or errors, the model needs to identify and correct errors in the table \\based on the user requirements, such as typos, duplicate values, or illegal characters and so on.\end{tabular} \\ 
\cline{2-3}
 & \begin{tabular}[c]{@{}c@{}}Data \\filtering\end{tabular} & \begin{tabular}[c]{@{}l@{}}Given a table and some filter criteria, the model is asked to filter some rows and columns in \\the table based on the given criteria. For example, only reserving rows that meet certain criteria.\end{tabular} \\ 
\cline{2-3}
 & \begin{tabular}[c]{@{}c@{}}Data \\classification\end{tabular} & \begin{tabular}[c]{@{}l@{}}Given a table and user requests, the model needs to classify table data into pre-defined categories. \\For example, classifying movie reviews in the given table into positive reviews or negative reviews.\end{tabular} \\ 
\cline{2-3}
 & Data sorting & \begin{tabular}[c]{@{}l@{}}The model needs to sort the data in the table according to the user's requirements and return\\~the sorted data, which can be sorted in the ascending or descending order.\end{tabular} \\ 
\hline
\multirow{2}{*}{\begin{tabular}[c]{@{}c@{}}Table \\Processing\end{tabular}} & \begin{tabular}[c]{@{}c@{}}Table \\modification\end{tabular} & \begin{tabular}[c]{@{}l@{}}Given the table and modification requirements, the model is asked to modify the overall table\\~according to the user's requirements and returns the processed table.\end{tabular} \\ 
\cline{2-3}
 & \begin{tabular}[c]{@{}c@{}}Format \\transformation\end{tabular} & \begin{tabular}[c]{@{}l@{}}The model needs to convert the original table to the desired format based on user requirements, such as \\from Markdown format to Latex format.\end{tabular} \\
\hline
\end{tabular}
}

\caption{Description of 20 seed tasks which are used to synthesize seed instructions based on synthetic tables.}
\label{seed_task_descriptions}
\end{table*}

\begin{table*}[t]\footnotesize
\centering
\renewcommand{\arraystretch}{1.3}
\setlength\tabcolsep{3pt}
\scalebox{0.8}{
\begin{tabular}{c|c|l} 
\hline
\textbf{Evolution Direction} & \textbf{Evolution Strategy} & \multicolumn{1}{c}{\textbf{Description}} \\ 
\hline
\multirow{7}{*}{Instruction Complication} & Add Constrains & adding one more constraints/requirements/conditions to the original instruction. \\ 
\cline{2-3}
 & Increase Depth & \begin{tabular}[c]{@{}l@{}}increasing the depth of the questions or requests in the original instruction. For instance, \\rewriting a simple question into a more profound question, or proposing a\\complex math problem instead of a simple calculation.\end{tabular} \\ 
\cline{2-3}
 & Add Reasoning Steps & \begin{tabular}[c]{@{}l@{}}increasing the required reasoning steps of the original instruction. For instance, if the original \\task can be solved with a few simple steps, you should rewrite it \\into more complex problems that request multi-step reasoning.\end{tabular} \\ 
\cline{2-3}
 & Add Task Number & \begin{tabular}[c]{@{}l@{}}adding more tasks/demands to the original instruction so that models need to perform multiple tasks. \\For instance, if the original instruction only contains one task, \\you can propose more tasks in the instruction and organize them in a Markdown list.\end{tabular} \\ 
\cline{2-3}
 & Add Details & replacing general concepts in the original instruction with more specific concepts. \\ 
\cline{2-3}
 & Increase Length & \begin{tabular}[c]{@{}l@{}}writing long and multi-line instructions. Each instruction consists of multiple lines \\or paragraphs of text to create complex tasks.\end{tabular} \\ 
\cline{2-3}
 & Add Context & \begin{tabular}[c]{@{}l@{}}designing more complex tasks which require not only the original input \\table but also additional input data (e.g., related contexts, code, \\background information or task examples, etc).\end{tabular} \\ 
\hline
\multirow{2}{*}{Instruction Generalization} & New Instruction & \begin{tabular}[c]{@{}l@{}}draw inspiration from the example tabular instruction and come up brand\\~new instructions about the provided table. New instructions require performing tasks \\that are different from example instructions.\end{tabular} \\ 
\cline{2-3}
 & Similar Instruction & \begin{tabular}[c]{@{}l@{}}come up with task instructions about the given table, which are similar with \\the original instruction. The new instructions SHOULD belong to the same \\task type or the same demand as the example instruction.\end{tabular} \\ 
\hline
\multirow{5}{*}{Table Generalization} & Change Format & convert the original table into a table in the new format \\ 
\cline{2-3}
 & Modify Header & \begin{tabular}[c]{@{}l@{}}paraphrasing some row headers or column headers into new headers \\with the same meaning. For instance, replacing original headers with synonyms.\end{tabular} \\ 
\cline{2-3}
 & Modify Data & \begin{tabular}[c]{@{}l@{}}replacing the data in the original table with new data. Make new data \\as diverse as possible. You can also replace some data with null values.\end{tabular} \\ 
\cline{2-3}
 & Order Permutation & randomly changing the order of rows and columns in the original table. \\ 
\cline{2-3}
 & Insert/Remove Data & randomly inserting or removing some new rows and columns. \\
\hline
\end{tabular}
}

\caption{Descriptions of 14 detailed data evolution strategies. In the evolution of each direction, one strategy is randomly sampled to fill in the corresponding data evolution prompt. }
\label{data_evol_strategies}
\end{table*}

\begin{table*}[t]\footnotesize
\centering
\renewcommand{\arraystretch}{1.3}
\setlength\tabcolsep{2pt}
\scalebox{0.9}{
\begin{tabular}{cccccccccccc} 
\hline
\textbf{Method} & \textbf{TABMWP} & \textbf{WTQ} & \textbf{HiTab} & \textbf{TabFact} & \textbf{InfoTabs} & \textbf{AIT-QA} & \textbf{TabMCQ} & \textbf{FeTaQA} & \textbf{QTSumm} & \textbf{TableGPT} & \textbf{Ave. Acc} \\ 
\hline
Llama3.1-8B-Instruct & 53.39 & 36.53 & 11.35 & 53.87 & 48.94 & 43.63 & 75.31 & 78.98 & 66.98 & 21.68 & 49.07 \\
w/ TableDreamer-27K & 60.57 & 42.47 & 17.25 & 57.32 & 49.98 & 56.75 & 82.99 & 84.67 & 75.12 & 33.03 & 56.02 \\
w/ TableDreamer-52K & 58.41 & 43.55 & 17.69 & 59.98 & 54.09 & 57.72 & 84.25 & 85.07 & 73.19 & 34.09 & 56.80 \\
$\triangle$ & -2.16 & 1.08 & 0.44 & 2.66 & 4.11 & 0.97 & 1.26 & 0.40 & -1.93 & 1.06 & 0.79 \\
\hline
\end{tabular}

}
\caption{The influence of adding 25K extra TableDreamer synthetic data (from Llama3.1-70B-Instruct).} 
\label{impact_of_adding_more_data}
\end{table*}

\begin{table*}[t]\footnotesize
\centering
\renewcommand{\arraystretch}{1.3}
\setlength\tabcolsep{2pt}
\scalebox{0.75}{
\begin{tabular}{c|c|c|ccccccc|c|cc|c} 
\hline
\multirow{2}{*}{\textbf{Exp. Setting}} & \multirow{2}{*}{\textbf{SFT Data}} & \multirow{2}{*}{\textbf{\# SFT Data}} & \multicolumn{8}{c|}{\textbf{Held-in }} & \multicolumn{2}{c|}{\textbf{Held-out }} & \multirow{2}{*}{\begin{tabular}[c]{@{}c@{}}\textbf{Overall }\\\textbf{Ave. Acc}\end{tabular}} \\ 
\cline{4-13}
 &  &  & \textbf{TABMWP} & \textbf{WTQ} & \textbf{HiTab} & \textbf{TabFact} & \textbf{InfoTabs} & \textbf{FeTaQA} & \textbf{QTSumm} & \textbf{Ave. Acc} & \textbf{AIT-QA} & \textbf{TabMCQ~} &  \\ 
\hline
\textbf{No Train Data} & Llama3.1-8B-Instruct & - & 53.39 & 36.53 & 11.35 & 53.87 & 48.94 & 78.98 & 66.98 & 50.01 & 43.63 & 75.31 & 52.11 \\ 
\hline
\begin{tabular}[c]{@{}c@{}}\textbf{Only }\\\textbf{Synthetic Data}\end{tabular} & w/ TableDreamer & 27K & 64.61 & 54.66 & 22.88 & 63.09 & 57.65 & 84.37 & \textbf{75.97} & 60.46 & 53.22 & 84.29 & 62.30 \\ 
\hline
\multirow{2}{*}{\begin{tabular}[c]{@{}c@{}}\textbf{A Few}\\\textbf{Human Data }\end{tabular}} & 50-shot & 350 & 56.18 & 37.75 & 14.78 & 56.34 & 47.88 & 83.23 & 69.48 & 52.23 & 51.07 & 77.84 & 54.95 \\ 
\cline{2-14}
 & w/ TableDreamer & 350+27K & 70.89 & 56.37 & 26.90 & 60.68 & 47.22 & 83.37 & 74.95 & 60.05 & \textbf{61.64} & 83.86 & 62.88 \\ 
\hline
\multirow{4}{*}{\begin{tabular}[c]{@{}c@{}}\textbf{All}\\\textbf{Human~Data }\end{tabular}} & Training Set Size & - & 30K & 17K & 8K & 31K & 18K & 8K & 5K & - & - & - & - \\ 
\cline{2-14}
 & Answer Type & - & CoT & S.A. & S.A. & S.A. & S.A. & L. A. & L. A. & - & - & - & - \\ 
\cline{2-14}
 & \begin{tabular}[c]{@{}c@{}}Ave.~\\Answer Length\end{tabular} & - & 65.30 & 1.73 & 1.40 & 1.03 & 1.35 & 18.60 & 50.60 & - & - & - & - \\ 
\cline{2-14}
 & All human data & 120K & 87.43 & 58.74 & 30.64 & 65.31 & 62.61 & 83.23 & 68.92 & 65.27 & 27.20 & 72.98 & 61.90 \\ 
\hline
\multicolumn{2}{c|}{\textbf{ $\triangle$ with only TableDreamer Data}} & - & {\cellcolor[rgb]{0.792,1,0.792}}22.82 & {\cellcolor[rgb]{0.792,1,0.792}}4.08 & {\cellcolor[rgb]{0.792,1,0.792}}7.76 & {\cellcolor[rgb]{0.792,1,0.792}}2.22 & {\cellcolor[rgb]{0.792,1,0.792}}4.96 & {\cellcolor[rgb]{1,0.792,0.792}}-1.14 & {\cellcolor[rgb]{1,0.792,0.792}}-7.05 & {\cellcolor[rgb]{0.792,1,0.792}}4.81 & {\cellcolor[rgb]{1,0.792,0.792}}-26.02 & {\cellcolor[rgb]{1,0.792,0.792}}-11.31 & {\cellcolor[rgb]{1,0.792,0.792}}-0.40 \\ 
\hline
\multicolumn{2}{c|}{\textbf{Human Data+\textbf{TableDreamer}~Data}} & 120K+27K & \textbf{91.90} & \textbf{62.27} & \textbf{31.02} & \textbf{76.56} & \textbf{71.87} & \textbf{84.22} & 73.19 & \textbf{69.98} & 32.87 & \textbf{85.61} & \textbf{67.59} \\ 
\hline
\multicolumn{2}{c|}{\textbf{ $\triangle$ with only Human Data}} & - & {\cellcolor[rgb]{0.792,1,0.792}}4.47 & {\cellcolor[rgb]{0.792,1,0.792}}3.53 & {\cellcolor[rgb]{0.792,1,0.792}}0.38 & {\cellcolor[rgb]{0.792,1,0.792}}11.25 & {\cellcolor[rgb]{0.792,1,0.792}}9.26 & {\cellcolor[rgb]{0.792,1,0.792}}0.99 & {\cellcolor[rgb]{0.792,1,0.792}}4.27 & {\cellcolor[rgb]{0.792,1,0.792}}4.71 & {\cellcolor[rgb]{0.792,1,0.792}}5.67 & {\cellcolor[rgb]{0.792,1,0.792}}12.63 & {\cellcolor[rgb]{0.792,1,0.792}}5.69 \\
\hline
\end{tabular}

}
\caption{Comparison of human-annotated data and TableDreamer-27K synthetic data (from GPT-4o) under different settings. `S.A.' and `L.A.' stand for `short answer' and `long answer' respectively. $\triangle$ indicates performance gap.} 
\label{combine_synthetic_and_human_data}
\end{table*}

\begin{table*}[t]\footnotesize
\centering
\renewcommand{\arraystretch}{1.3}
\setlength\tabcolsep{2pt}
\scalebox{0.87}{
\begin{tabular}{c|cc|cc|c|cccc|c} 
\hline
\multirow{3}{*}{\textbf{Method}} & \multicolumn{5}{c|}{\textbf{IFEval}} & \multicolumn{5}{c}{\textbf{MMLU }} \\ 
\cline{2-11}
 & \multicolumn{2}{c}{\textbf{prompt-level}} & \multicolumn{2}{c|}{\textbf{instruction-level}} & \multirow{2}{*}{\textbf{Ave. Acc.}} & \multicolumn{1}{l}{\multirow{2}{*}{\textbf{Humanities}}} & \multicolumn{1}{l}{\multirow{2}{*}{\textbf{Social Science}}} & \multicolumn{1}{l}{\multirow{2}{*}{\textbf{STEM}}} & \multicolumn{1}{l|}{\multirow{2}{*}{\textbf{Other}}} & \multicolumn{1}{l}{\multirow{2}{*}{\textbf{Ave Acc.}}} \\ 
\cline{2-5}
 & \textbf{~Strict Acc.} & \textbf{Loose Acc.} & \textbf{Strict Acc.} & \textbf{Loose Acc.} &  & \multicolumn{1}{l}{} & \multicolumn{1}{l}{} & \multicolumn{1}{l}{} & \multicolumn{1}{l|}{} & \multicolumn{1}{l}{} \\ 
\hline
Qwen2.5-7B-Instruct & 71.34 & 73.38 & 79.37 & 80.93 & 76.26 & 64.59 & 82.22 & 81.01 & 77.61 & 74.99 \\ 
\hline
\begin{tabular}[c]{@{}c@{}}TableGPT2 \\(Qwen2.5-7B)\end{tabular} & 52.86 & 57.67 & 62.82 & 67.14 & 60.12 & 59.49 & 78.55 & 75.08 & 74.43 & 70.47 \\
$\triangle$ & {\cellcolor[rgb]{1,0.69,0.69}}18.48 & {\cellcolor[rgb]{1,0.69,0.69}}15.71 & {\cellcolor[rgb]{1,0.69,0.69}}16.55 & {\cellcolor[rgb]{1,0.69,0.69}}13.79 & {\cellcolor[rgb]{1,0.69,0.69}}16.13 & {\cellcolor[rgb]{1,0.69,0.69}}5.10 & {\cellcolor[rgb]{1,0.69,0.69}}3.67 & {\cellcolor[rgb]{1,0.69,0.69}}5.93 & {\cellcolor[rgb]{1,0.69,0.69}}3.18 & {\cellcolor[rgb]{1,0.69,0.69}}4.52 \\ 
\hline
Llama3.1-8B-Instruct & 68.94 & 73.19 & 76.73 & 80.45 & 74.83 & 61.11 & 75.76 & 69.52 & 75.35 & 69.41 \\ 
\hline
\begin{tabular}[c]{@{}c@{}}TableBenchLLM \\(Llama3.1-8B)\end{tabular} & 20.88 & 25.13 & 31.65 & 36.45 & 28.53 & 15.75 & 22.59 & 12.62 & 11.29 & 15.55 \\
$\triangle$ & {\cellcolor[rgb]{1,0.69,0.69}}48.06 & {\cellcolor[rgb]{1,0.69,0.69}}48.06 & {\cellcolor[rgb]{1,0.69,0.69}}45.08 & {\cellcolor[rgb]{1,0.69,0.69}}44.00 & {\cellcolor[rgb]{1,0.69,0.69}}46.30 & {\cellcolor[rgb]{1,0.69,0.69}}45.36 & {\cellcolor[rgb]{1,0.69,0.69}}53.17 & {\cellcolor[rgb]{1,0.69,0.69}}56.90 & {\cellcolor[rgb]{1,0.69,0.69}}64.06 & {\cellcolor[rgb]{1,0.69,0.69}}53.86 \\ 
\hline
\begin{tabular}[c]{@{}c@{}}TableLLM \\(CodeLlama-7B)\end{tabular} & 18.66 & 22.92 & 28.65 & 32.61 & 25.71 & 12.55 & 16.83 & 16.29 & 16.09 & 17.30 \\
$\triangle$ & {\cellcolor[rgb]{1,0.69,0.69}}50.28 & {\cellcolor[rgb]{1,0.69,0.69}}50.27 & {\cellcolor[rgb]{1,0.69,0.69}}48.08 & {\cellcolor[rgb]{1,0.69,0.69}}47.84 & {\cellcolor[rgb]{1,0.69,0.69}}49.12 & {\cellcolor[rgb]{1,0.69,0.69}}48.56 & {\cellcolor[rgb]{1,0.69,0.69}}58.93 & {\cellcolor[rgb]{1,0.69,0.69}}53.23 & {\cellcolor[rgb]{1,0.69,0.69}}59.26 & {\cellcolor[rgb]{1,0.69,0.69}}52.11 \\ 
\hline
\begin{tabular}[c]{@{}c@{}}TableDreamer-27K \\(ours)\end{tabular} & 68.20 & 71.90 & 76.61 & 79.73 & 74.11 & 61.42 & 75.43 & 71.24 & 74.95 & 69.73 \\
$\triangle$ & {\cellcolor[rgb]{1,0.894,0.894}}0.74 & {\cellcolor[rgb]{1,0.894,0.894}}1.29 & {\cellcolor[rgb]{1,0.894,0.894}}0.12 & {\cellcolor[rgb]{1,0.894,0.894}}0.72 & {\cellcolor[rgb]{1,0.894,0.894}}0.72 & {\cellcolor[rgb]{0.867,1,0.867}}+0.31 & {\cellcolor[rgb]{1,0.894,0.894}}0.33 & {\cellcolor[rgb]{0.867,1,0.867}}+1.72 & {\cellcolor[rgb]{1,0.894,0.894}}0.40 & {\cellcolor[rgb]{0.867,1,0.867}}+0.32 \\
\hline
\end{tabular}

}
\caption{Comparison of TableDreamer-27K and existing tabular LLMs on IFEval and MMLU benchmarks. $\triangle$ indicates performance decrease of different tabular LLMs compared with general LLMs in the same/similar-series.} 
\label{general_capacity_results}
\end{table*}

\section{More Results and Analysis}
\label{sec:more_results_and_analysis}

\subsection{Effect on Different LLMs} 
As shown in Table \ref{different_LLM_results}, other LLMs can also benefit from fine-tuning with TableDreamer-27K data, indicating the transferability of synthetic data. Compared with Llama3.1-8B-Instruct, the performance gains of three LLMs are relatively smaller, which may be because we used Llama3.1-8B-Instruct as the target LLM to identify vulnerability data in order to achieve targeted performance enhancement.

To more thoroughly investigate the generalization of weakness detection beyond Llama3.1-8B-instruct, we conduct extra ablation experiments by using the TableDreamer-34K data (from GPT-4o) without weakness data selection to fine-tune other LLMs. Performance comparison is shown in Table \ref{different_LLM_results_without_weakness_idtentification} and the '$\triangle$' indicates the performance gap between the normal TableDreamer-27K and unselected 34K data. We can observe that, compared with 27K filtered data, using 34K unfiltered data also leads to performance decrease for other LLMs, but their gap is smaller than that of Llama3.1-8B-instruct. For example, the average accuracy gap on InternLM2.5-7B-Chat is 2.57 and the gap on MiniCPM3-4B is only 0.78. This demonstrates that the detected weakness data with Llama3.1-8B-Instruct can generalize to other LLMs but with different extent, which could be attributed to the behavior similarity between different LLMs resulting from model distillation~\cite{quantify_model_distillation}. Intuitively, if the target LLM used for weakness detection and another LLM both utilize fine-tuning data distilled from the same teacher LLM (like GPT-4o), their model behavior could be similar or homogeneous, thus they are likely to share some weakness data in table understanding.

\subsection{Credibility of Weakness Detection} The credibility of LLM-as-a-judge-based weakness detection module directly impacts on the reserved synthetic training data. Thus, we conduct an extra ablation experiment by randomly selecting weakness data in each round of data evolution and use Llama3.1-70B-Instruct to synthesize 27K data for fine-tuning Llama3.1-8B-Instruct. From the results in Table \ref{ablation_study_of_randomly_selected_weakness_data}, we can observe that randomly selecting weakness data leads to a substantial degeneration of 3.29 in average accuracy over 10 benchmarks, which validates the effectiveness of the LLM-as-a-judge-based weakness detection. Besides, with more complex task instructions being synthesized, we argue that more advanced methods need to be adopted to provide reliable LLM-based evaluation for selecting weakness data, e.g., criteria decomposition or majority voting~\cite{llmasajudge_survey_1,llmasajudge_survey_2}.

\subsection{Combining Synthetic and Human Data} Although the results in Table \ref{main_results} and Table \ref{low_resource_results} have shown that TableDreamer synthetic data could improve table understanding ability under zero-shot (no human-annotated data) and few-shot (limited amount of human-annotated data) scenarios, we want to further investigate whether our synthetic data can effectively complement human-annotated training data in tabular benchmarks and compare the quality of human-annotated and synthetic data. To this end, we fine-tune Llama3.1-8B-Instruct with all training data of 7 benchmarks and also combine these human-annotated data with synthetic TableDreamer-27K data. We leave AIT-QA and TABMCQ as held-out benchmarks and do not use their training data.

The results in Table \ref{combine_synthetic_and_human_data} reveal that, using all human-annotated training data indeed greatly improves performance on held-in benchmarks, but at the significant cost of held-out performance. Compared with only using 27K TableDreamer data, using 120K human-annotated data boosts the held-in average accuracy from 60.46 to 65.27, but the held-out performance  greatly degenerates, e.g., the performance on AIT-QA declines from 53.22 to 27.20, which eventually leads to a worse average performance over 9 benchmarks (61.90 vs 62.30). We believe this performance instability is due to the different quality of human-labeled responses. As shown in the 'Answer Type' row, TABMWP contains high-quality human-annotated responses with chain-of-thoughts, thus leading to the largest accuracy boost of 22.82, but most human-annotated data only contains short answers, which are very prone to overfitting on dataset-specific patterns and harm out-of-distribution and general capacities.

By contrast, the synthetic TableDreamer data could act as a supplementary part, which not only provides better diversity of tables, instructions, and tasks, but also includes detail responses from teacher LLMs, resulting in substantial performance boost under both few-shot and standard training settings. For instance, using both human-annotated and TableDreamer data obtains the best average performance of 67.59. Moreover, human-generated data also faces challenges like high costs and lacking creativity in synthesizing diverse table-related instructions and tasks, where LLM-generated data could serve as a viable alternative or supplement~\cite{llm_based_data_generation_survey}.

\subsection{General Capacity of Tabular LLMs}
It is very important for tabular LLMs to maintain their general ability on non-tabular tasks such as instruction-following or commonsense question answering. As a result, we evaluate our method and existing tabular LLMs on two general LLM benchmarks IFEval~\cite{IFEval_benchmark} and MMLU~\cite{MMLU}. The IFEval is an instruction following benchmark which assesses LLMs' ability to follow natural language instructions, e.g., `Write a casual summary of LLMs with two sections and at least 25 sentences'. MMLU is a multi-task benchmark where LLMs needs to answer multi-choice questions from 57 subjects such as elementary mathematics and computer science.

For IFEval, we follow the original paper and report prompt-level and instruction-level accuracy under the strict and loose settings (i.e., 4 metrics), which represents the percentage of prompts and instructions that LLMs successfully followed. For MMLU, we report exact match accuracy of four broad disciplines: Humanities, Social Science, STEM and Other. The zero-shot CoT setting is used for both benchmarks. For MMLU, we add requirements in the input prompt and ask LLMs to represent the final answer in the JSON format for answer parsing and accuracy computation.

The results in Table \ref{general_capacity_results} reveal that existing tabular LLMs suffer tremendous performance drop on general benchmarks, e.g., compared with Llama3.1-8B-Instruct, the TableBenchLLM only achieves average accuracy of 28.53 on IFEval (46.30$\downarrow$) and 15.55 on MMLU (53.86$\downarrow$). The average performance of powerful TableGPT2 also declines substantially by 16.13 on IFEval and 4.52 on MMLU. These phenomena correspond to our findings and findings in~\citet{rethinking_table_instruction_tuning}, i.e., existing tabular LLMs can only perform well under the in-distribution table understanding setting and they significantly sacrifice out-of-distribution as well as fundamental general capabilities.

In comparison, our method maintains general capabilities with slight performance fluctuations on IFEval (-0.72) and MMLU (+0.32), which validates that our method can effectively enhance table understanding performance without sacrificing broader and general capabilities.

\subsection{Effect of Adding More Data}

Another important question is whether adding more data would continue to improve model performance or if the gains would eventually plateau. Thus, considering the high cost of GPT-4o API, we utilize Llama3.1-70B-instruct to synthesize 25K new data and combine it with the original 27K Llama3.1-70B-instruct-generated data to fine-tune Llama3.1-8B-instruct. The results are listed in Table \ref{impact_of_adding_more_data} and '$\triangle$' indicates the performance increase from 27K data to 52K data. We can find that, although increasing the synthetic data volume improves the average performance, the improvement is much more smaller compared to the accuracy boost brought by 27K data. We anticipate that the performance gains will plateau with more synthetic data. For one thing, as we continually explore the input space by generating more complex instructions, it could reach the capability boundary of the teacher LLM, i.e., the synthetic table-related instructions are beyond the capacity of Llama3.1-70B-instruct, which will lead to more problematic responses that bring noise and negative effect to model training. For another, it would also become more difficult for the LLM-as-a-judge to rate the correctness of student LLMs' responses of more complicated tasks, resulting in potentially unreliable weakness data. Therefore, we think it would be an intriguing idea to introduce a monitoring module within the Strong2Weak data synthesis approaches, which determines when to stop the data synthesis process rather than endlessly distilling new data from the stronger LLM.

\subsection{Case Study.} 

\begin{figure*}[t]
  \centering
  \includegraphics[width=\linewidth]{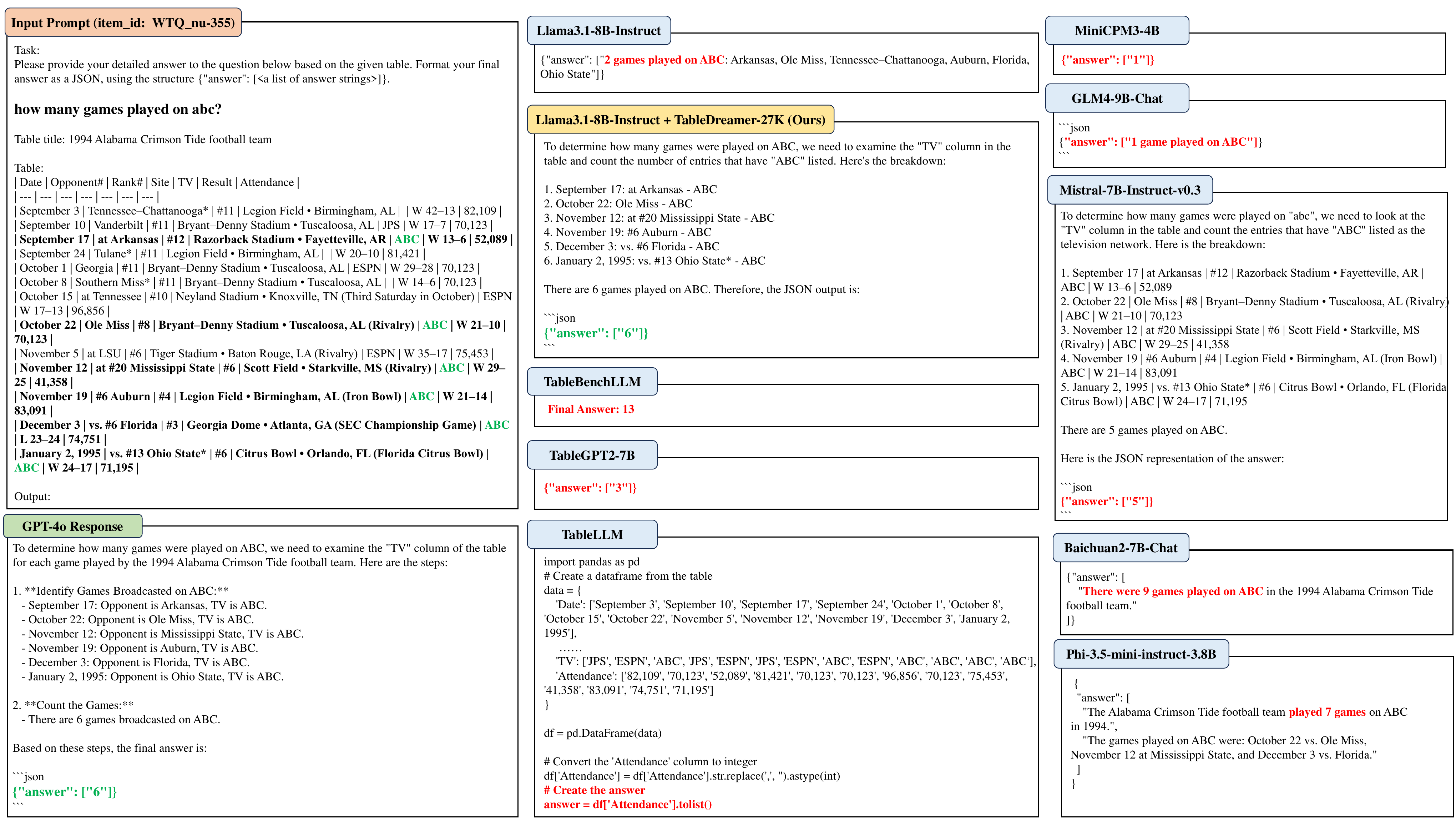}
  \caption{Qualitative comparison of model responses on WTQ benchmark.}
  \label{case_study_1}
\end{figure*}

\begin{figure*}[t]
  \centering
  \includegraphics[width=\linewidth]{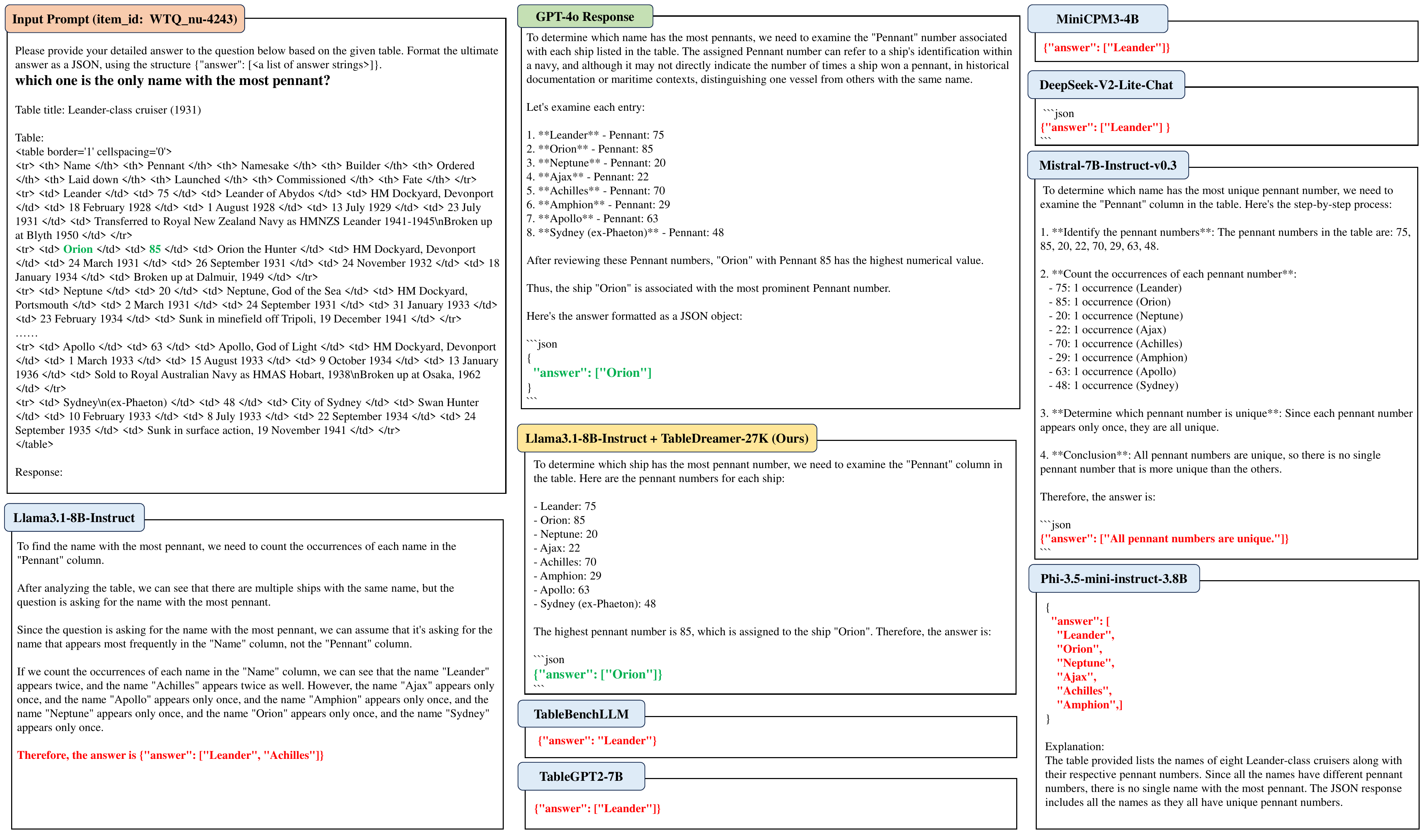}
  \caption{Qualitative comparison of model responses on WTQ benchmark.}
  \label{case_study_2}
\end{figure*}

\begin{figure*}[t]
  \centering
  \includegraphics[width=\linewidth]{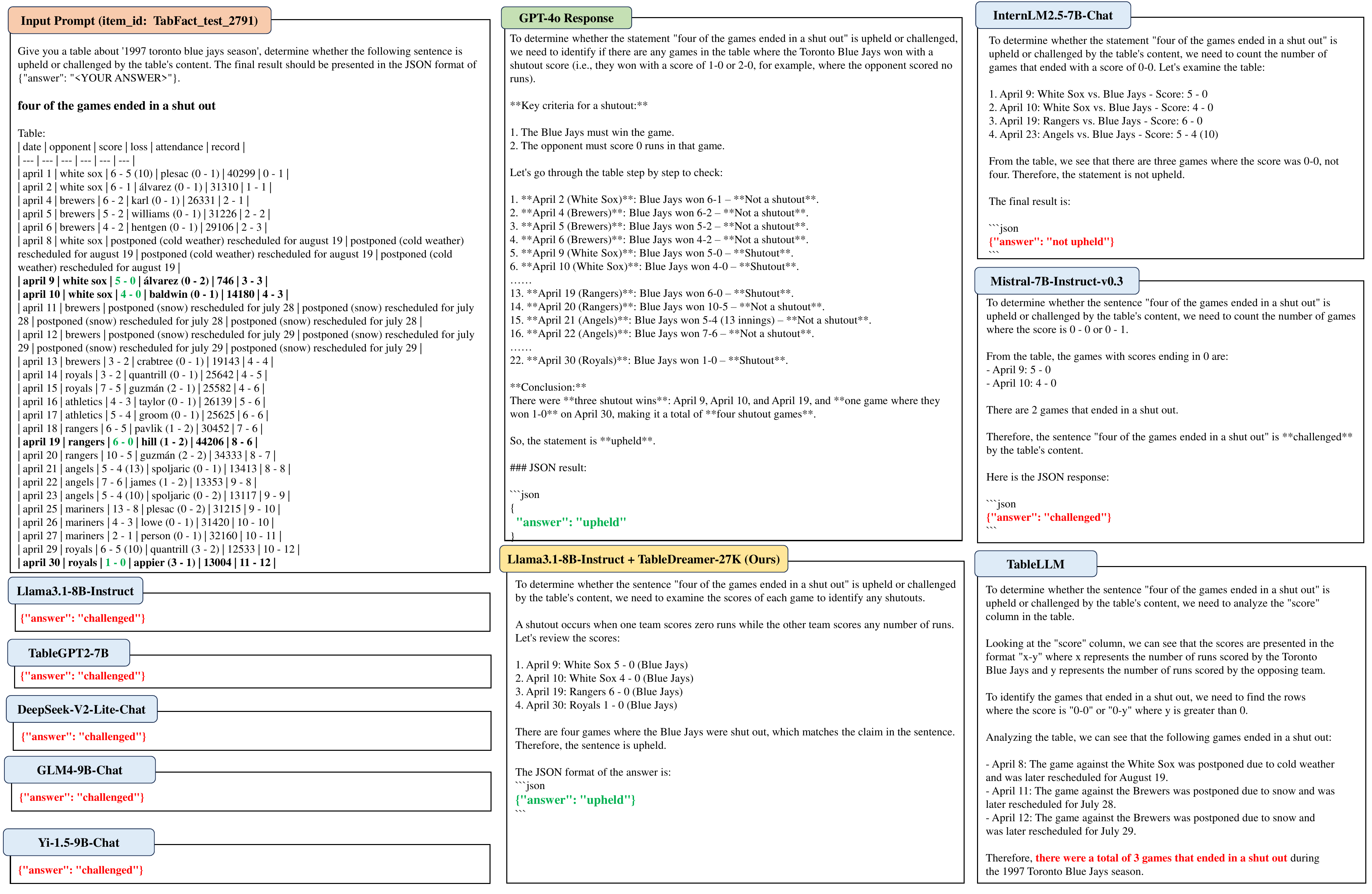}
  \caption{Qualitative comparison of model responses on TabFact benchmark.}
  \label{case_study_3}
\end{figure*}

\begin{figure*}[t]
  \centering
  \includegraphics[width=\linewidth]{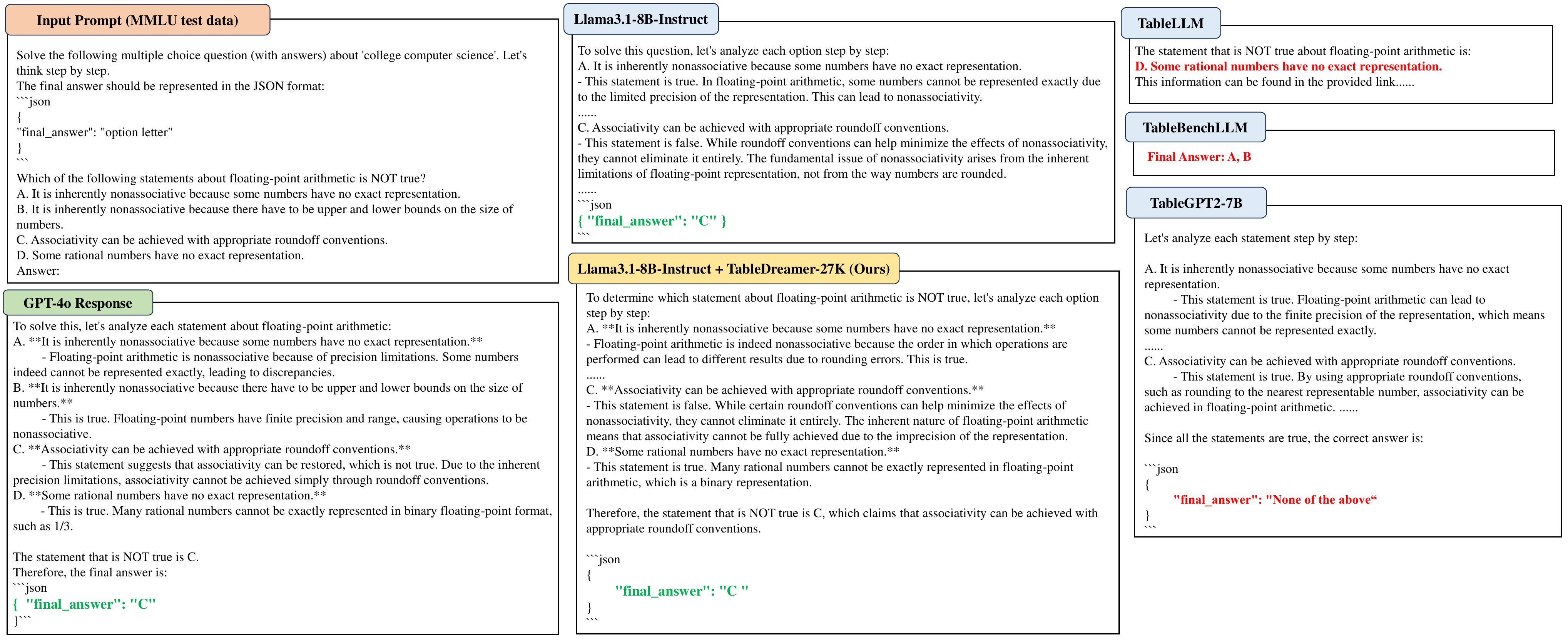}
  \caption{Qualitative comparison of model responses on MMLU benchmark.}
  \label{case_study_4}
\end{figure*}

We conduct a side-by-side qualitative comparison of TableDreamer with other baselines, as illustrated in Figure \ref{case_study_1}-\ref{case_study_4}. The results demonstrate that TableDreamer synthetic data can improve the table understanding ability of vanilla Llama3.1-8B-Instruct and outperform strong baselines including recent tabular LLMs. Moreover, case study on the MMLU benchmark in Figure \ref{case_study_4} also intuitively shows that TableDreamer does not sacrifice the general capacity of LLM, while other tabular LLMs such as TableLLM, TableBenchLLM and even TableGPT2 suffer significant decline of their general abilities, which underscores the importance of diverse table instruction-tuning data to avoid ovefitting.

\begin{figure*}[t]
  \centering
  \includegraphics[width=0.8\linewidth]{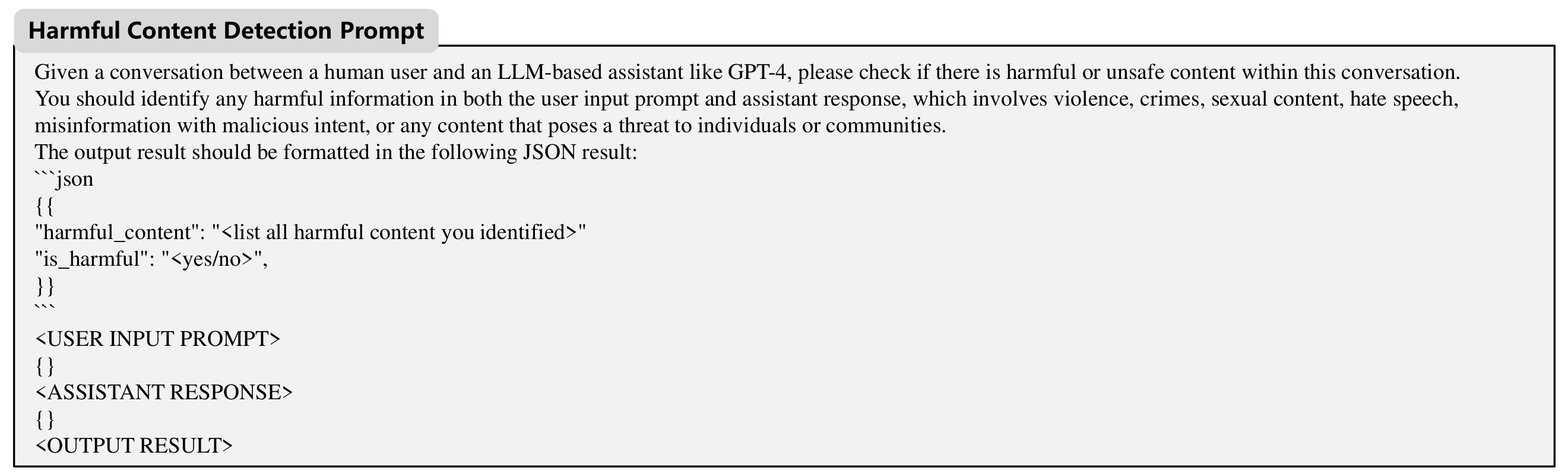}
  \caption{The prompt used for harmful content detection.}
  \label{harmful_content_detection_prompt}
\end{figure*}

\end{document}